\documentclass[5p, twocolumn, final]{elsarticle}
\newif\ifRowPlots
\RowPlotsfalse

\newif\ifarxive
\arxivetrue

\newif\ifTikzScatterPlots
\TikzScatterPlotstrue

\usepackage{lineno,hyperref}

\usepackage{amsmath}
\usepackage{amssymb}

\usepackage{morefloats}

\usepackage{adjustbox}

\usepackage{epsfig} 
\usepackage{graphicx}
\graphicspath{{./figs/}}
\usepackage[tight,normalsize,sf,SF]{subfigure}

\usepackage{pgfplots} 
\usepackage{soul}
\usepackage{rotating}

\usepackage{pgf,tikz}
\usetikzlibrary{spy,backgrounds,plotmarks}
\pgfplotsset{compat=newest}

\tikzset{every picture/.style={font issue=\scriptsize},
         font issue/.style={execute at begin picture={#1\selectfont}}
        }

\usetikzlibrary{external}
\newcommand{\includetikz}[1]{%
    \includegraphics{figs/{#1.tikz}.pdf}%
}

\usepackage[noend,linesnumbered]{algorithm2e} 

\usepackage{booktabs}
\usepackage{multirow}
\usepackage{dblfloatfix}    %

\usepackage{afterpage}
\usepackage{enumitem}

\DeclareMathOperator{\diag}{diag}

\DeclareMathOperator{\VEC}{vec}

\newcommand{\R}{\mathbb{R}}

\newcommand{\onevec}{\mathbf{1}}
\newcommand{\zerovec}{\mathbf{0}}
\newcommand{\matI}{\mathbf{I}}
\newcommand{\matX}{\mathbf{X}}
\newcommand{\matP}{\mathbf{P}}

\newcommand{\matPhi}{\mathbf{\Phi}}

\newcommand{\prob}{\mathfrak{p}}

\newcommand{\matW}{\mathbf{W}}

\newcommand{\matSigma}{\mathbf{\Sigma}}

\newcommand{\matA}{\mathbf{A}}

\newcommand{\vecb}{\mathbf{b}} 

\newcommand{\vecalpha}{\boldsymbol\alpha}

\newcommand{\vecp}{\mathbf{p}}

\newcommand{\vectheta}{\boldsymbol{\theta}}

\newcommand{\vecx}{\mathbf{x}}
\newcommand{\vecy}{\mathbf{y}}

\newcommand{\matC}{\mathbf{C}}

\biboptions{authoryear}

\ifarxive
  \makeatletter
  \def\ps@pprintTitle{%
    \let\@oddhead\@empty
    \let\@evenhead\@empty
    \def\@oddfoot{\reset@font\hfil\thepage\hfil}
    \let\@evenfoot\@oddfoot
  }
  \makeatother
\else
  \journal{Medical Image Analysis}
  \modulolinenumbers[5]
\fi 

\makeatletter
\def\@author#1{\g@addto@macro\elsauthors{\normalsize%
    \def\baselinestretch{1}%
    \upshape\authorsep#1\unskip\textsuperscript{%
      \ifx\@fnmark\@empty\else\unskip\sep\@fnmark\let\sep=,\fi
      \ifx\@corref\@empty\else\unskip\sep\@corref\let\sep=,\fi
      }%
    \def\authorsep{\unskip,\space}%
    \global\let\@fnmark\@empty
    \global\let\@corref\@empty  %
    \global\let\sep\@empty}%
    \@eadauthor={#1}
}
\makeatother

\begin{document}

\begin{frontmatter}

\title{Shape-aware Surface Reconstruction from Sparse 3D Point-Clouds}

\author{Florian Bernard\corref{corrauthor}\fnref{chl,lcsb}}
\cortext[corrauthor]{Corresponding author}
\author{Luis Salamanca\fnref{lcsb}}
\author{Johan Thunberg\fnref{lcsb}}
\author{Alexander Tack\fnref{zib}}
\author{Dennis Jentsch\fnref{zib}}
\author{Hans Lamecker\fnref{zib,ts}}
\author{Stefan Zachow\fnref{zib,ts}}
\author{Frank Hertel\fnref{chl}} 
\author{Jorge Goncalves\fnref{lcsb}}
\author{Peter Gemmar\fnref{lcsb,tuas}}

\fntext[chl]{Centre Hospitalier de Luxembourg, Luxembourg}
\fntext[lcsb]{Luxembourg Centre for Systems Biomedicine, University of Luxembourg, Esch-sur-Alzette, Luxembourg}
\fntext[zib]{Zuse Institute Berlin (ZIB), Germany}
\fntext[ts]{1000shapes GmbH, Berlin, Germany}
\fntext[tuas]{Trier University of Applied Sciences, Trier, Germany}

\begin{abstract}
The reconstruction of an object's shape or surface from a set of 3D points plays an important role in medical image analysis, e.g. in anatomy reconstruction from tomographic measurements or in the process of aligning intra-operative navigation and preoperative planning data. In such scenarios, one usually has to deal with \emph{sparse} data, which significantly aggravates the problem of reconstruction. However, medical applications often provide contextual information about the 3D point data that allow to incorporate prior knowledge about the shape that is to be reconstructed. To this end, we propose the use of a statistical shape model (SSM) as a prior for surface reconstruction. The SSM is represented by a point distribution model (PDM), which is associated with a surface mesh. Using the shape distribution that is modelled by the PDM, we formulate the problem of surface reconstruction from a probabilistic perspective based on a Gaussian Mixture Model (GMM). In order to do so, the given points are interpreted as samples of the GMM. By using mixture components with anisotropic covariances that are ``oriented'' according to the surface normals at the PDM points, a surface-based fitting is accomplished. Estimating the parameters of the GMM in a maximum a posteriori manner yields the reconstruction of the surface from the given data points. We compare our method to the extensively used Iterative Closest Points method on several different anatomical datasets/SSMs (brain, femur, tibia, hip, liver) and demonstrate superior accuracy and robustness on sparse data.
\end{abstract}


\end{frontmatter}


\section{Introduction}
The reconstruction of an object's shape or surface from a set of 3D points is a highly relevant problem in medical image analysis. It appears for example in image segmentation, where images provide implicit information on the location of anatomical structures via intensity levels, which is frequently converted into geometric information via some kind of feature extraction method.
  Another scenario is computer-assisted surgery, where a pre-operative therapy plan is transferred to the operating room by means of a navigation system.

In contrast to mere 3D point-clouds that may represent virtually any object, image data of medical objects yield additional contextual information that can be used to adopt prior knowledge about the anatomical structures to be reconstructed. \citet{Heckel:2011jk} make use of the variational interpolation method \citep{Turk:2005vx}, which essentially uses a general prior on the surface smoothness. 
Going one step further, \citet{Pauly:2005wb} and \citet{Gal:2007fk} have considered templates for 3D scan completion that are matched to measurements. However, these methods are limited since the available measurements are assumed to be sufficiently dense \citep{Berger:2014gn}.
To tackle this limitation, \citet{Bernard:2015wx} suggested the use of a statistical shape model (SSM) for surface reconstruction. 
Because the class of anatomical structures is known for clinical routine tasks such as segmentation, registration, or intra-operative navigation, it is possible to use their shapes as geometric priors.

\label{keycontributions}
Our main contribution of this work is the introduction of a \emph{surface-based} SSM fitting procedure in order to reconstruct a surface from a sparse 3D point-cloud.
Using a point distribution model (PDM) \citep{Cootes:1992uw}, we incorporate a prior into our reconstruction framework that captures the likely shape of the object to be extracted. In doing so, we reformulate the problem of surface reconstruction from a probabilistic perspective, embedding the prior distribution of the SSM parameters that explain plausible shapes into the objective function. 
Our evaluation considers several different anatomical structures and SSMs (brain, femur, tibia, hip, liver) and sparse data point scenarios, which may occur in different applications, ranging from interactive segmentation to intra-operative registration for navigation. We are able to show superior accuracy and robustness compared to the extensively used Iterative Closest Point (ICP) method \citep{Besl:1992iv}. Rather than restricting ourselves to particular applications by investigating application-specific aspects, our goal is to demonstrate the \emph{general applicability} of the proposed approach in order to emphasize that the method may be useful in a wide range of settings.
The surface-based fitting procedure is achieved by the following methodological contributions:
\begin{itemize}
  \item By extending existing point-set registration procedures based on Gaussian Mixture Models (GMMs) \citep{Myronenko:ve,Myronenko:2010wn,Horaud:2011kz,Zheng:2013ue} such that anisotropic covariances are used in combination with a PDM as transformation model, we obtain a shape-aware surface reconstruction method that is superior to ICP with respect to robustness and accuracy.
  \item Before, only spherical (isotropic) GMMs accounting for a \emph{point-based} matching have been used \citep{Zheng:2013ue,Bernard:2015wx}. We now complement these works by presenting a formulation that is based on anisotropic GMMs that are ``oriented'' by the surface normals, accounting for a \emph{surface-based} fitting. 
  \item A \emph{rigorous and self-contained derivation} of the surface-based fitting method is presented, leading to an Expected Conditional Maximisation (ECM) algorithm \citep{Meng:1993uq}. ECM shares the same convergence properties as the Expectation Maximisation (EM) method \citep{Dempster:1977ul} while being more general. 
  \item We develop a fast approximation of the ECM-based fitting method that has the same computational complexity as the spherical GMM-based method. Numerical simulations show that it is less prone to unwanted local optima compared to the original ECM-based method. %
\end{itemize}

This article is organised as follows: section~\ref{relatedWork} summarises previous research relevant to our methodology. In sections~\ref{background} and~\ref{probspec}, we introduce our notation and formally state the considered surface reconstruction problem. In addition, we recapitulate the background of PDMs, probabilistic point-set registration, and the Expectation Maximisation method.
In section~\ref{methods} we present our novel shape-aware surface reconstruction method, including a time complexity and convergence analysis. Section~\ref{experiments} comprises experiments conducted using the proposed methods. In section~\ref{conclusion} we conclude this work.

\section{Related Work}\label{relatedWork}

A plethora of methods for general surface reconstruction has been presented in the literature so far (see e.g. \citet{Raya:1990fv,Bolle:1991fg,Herman:1992co,Hoppe:1992im,Edelsbrunner:1994vn,Bajaj:1995tv,Amenta:1998wv,Bernardini:1999iu,Treece:2000ug,Kazhdan:2006vg,Schroers:2014tn}). Many of them are summarised and described in the State-of-the-Art Report (STAR) by \citet{Berger:2014gn}. In the remainder of the present section, we will discuss only those surface reconstruction methods that go beyond pure smoothness assumptions and make use of more explicit shape priors.

For the completion of 2D shapes, \citet{Guo:2012gg,Guo:2013fe} incorporate templates from a shape database as (geometric) priors into a Bayesian framework. 
Similarly, a database of 3D shapes is used by \citet{Pauly:2005wb} for completing 3D surface scans. 
For increased flexibility compared to static priors, \citet{Gal:2007fk} use a context-specific database of \emph{local} shape priors, where the input data is matched by (dynamically) combining multiple local shape priors into a global prior. 
As pointed out by \citet{Berger:2014gn}, both approaches described by \citet{Pauly:2005wb} and \citet{Gal:2007fk} are limited by the assumption that the point-clouds are assumed to be sufficiently dense. 
A unified framework for repairing the geometry and texture of meshes has been presented by \citet{Park:2006fh}. They employ context-based geometry filling for filling holes in the surface, where available local patches of the mesh are used to fill its missing parts. 
The task of obtaining high-resolution 3D meshes from low-quality inputs is tackled by \citet{Shen:2012gg} by dynamically assembling object templates from a database of object parts. 
The 3D shape completion methods discussed above have a common focus on completing (mostly small) missing parts of meshes obtained from 3D scans. However, our interest lies primarily in methods that go beyond patching or improving low-resolution input.

\citet{Blanz:2004cz} have presented a closed-form solution for SSM-based 3D surface reconstruction from a sparse set of points, which relies, however, on the assumption that such a set of points is already in correspondence to the model. \citet{Albrecht:2013tw} introduced posterior shape models that have the objective to model the distribution of a whole shape given only partial information. This method assumes that the corresponding model points of the available partial data are known. In their experiments this issue is either solved manually or using the ICP method. Similarly, for shape prediction from sparse observations, \citet{Blanc:2012jv} use a variant of ICP that evaluates multiple initialisations. 
\citet{Anguelov:2005wy} introduced the shape completion and animation of people (SCAPE) method, where one model for pose deformations and one model for shape variations are learned separately. The main objective of this method is the completion of body shapes based on a small number of known positions for some of the model points. 
Applied to bone models, \citet{Rajamani:2007vx} fit an SSM to a small number of anatomical landmarks that correspond to some of the model points. 
\citet{Baka:2010tv} fit an SSM to sparse data points that are in correspondence with the model, applied to 2D heart datasets.
By producing confidence intervals as output, their method is able to incorporate uncertainties in the input data. 
Instead of using a trained SSM, \citet{Lu:2011vb}  formulate a low-rank matrix recovery problem for restoring missing parts of objects in archaeological studies. Considering that a set of (incomplete) objects of the same class is available, and that correspondences between common parts are known, their approach is based on the assumption that all shapes are approximately linearly correlated. 
A shortcoming of the methods discussed so far is that they all \emph{assume known correspondences.} However, if the objects do not exhibit a sufficient amount of distinct features, the identification of exact correspondences is very difficult or even infeasible in practice.

In the literature, there are several methods published addressing this difficulty in (automatically) detecting the correspondence between sparse data points and a model. 
Due to its simplicity, the ICP algorithm, where correspondences and transformations are estimated in an alternating manner, is a very popular method for the registration of two shapes represented as point sets. Numerous variants of the originally-proposed method have since been developed, e.g. by \citet{Rusinkiewicz:2001wp,Granger:2002wb,Segal:2009ws,MaierHein:2012wo,Bouaziz:2013ue,Billings:2015wb}.
For example, \citet{Granger:2002wb} propose the EM-ICP algorithm where hidden variables are used to model unknown correspondences in a surface registration problem. Based on this work, \citet{Hufnagel:2008bg} have proposed a method for learning a PDM from unstructured point-sets. For that, the authors establish probabilistic correspondences first, followed by the computation of the mean shape and the modes of variability.
The surface reconstruction method by \citet{Stoll:2006fb} is able to deform a given template to fit point-cloud data. To do so, the user defines initial correspondences between the template and a set of points, which are then refined iteratively in an ICP-like manner.
Along the lines of \citet{Stoll:2006fb}, for knee surgery, \citet{Fleute:2006eo,Fleute:1999wb} have presented a methodology to find pose and shape deformation parameters in order to fit an SSM to very sparse data points. Their approach resembles ICP due to the alternating closest point estimation and pose/deformation model parameter updates. 
Similarly, \citet{Chan:2003tz} reconstruct 3D models of bones in orthopaedic surgery based on a sparse set of points obtained from ultrasound. Here, shapes are repeatedly instantiated using a PDM, where each shape instance is used as input to ICP in order to (rigidly) fit the sparse points. This work has been extended by \citet{Barratt:2008tz}, who use a PDM defined on a regular grid instead of the shape surface.
\citet{Zheng:2006wp} have proposed a three-stage procedure for sparse shape reconstruction in computer-assisted orthopaedic surgery. In the first stage, the pose of the sparse points is adapted using ICP so that they best fit the mean shape. In the second stage, the shape deformation parameters are estimated for the given correspondences, which are eventually refined using a further deformation based on thin-plate splines.
Based on Gaussian Mixture Models with heteroscedastic covariances, \citet{Zheng:2013ue} has proposed a method for deformable shape registration.
A promising approach of fitting a hand pose model to points that are sufficiently densely sampled from depth images is presented by \citet{Taylor:2014bl,Taylor:2016aa}. The authors formulate a continuous optimisation problem that solves simultaneously for surface-based correspondences and model parameters using a nonlinear sum-of-squares objective.

In the work of \citet{Chang:2009ex}, the registration of articulated shapes in two range scans is based on a reduced deformation model defined on a regular grid. In this approach, the deformations are modelled by a convex combination of rigid transformations, where the weights are spatially varying. The registration is performed by alternatingly updating the rigid transformations and their weights, where closest point correspondences are recomputed after each step.

Despite the immense amount of literature on applications and variations of statistical shape models (for an overview the interested reader is referred to the works by \citet{Cootes:1992uw,Ginneken02activeshape,Heimann:2009kv}), there have been a limited number of investigations into the use of SSMs for interactive segmentation.
In the work of \citet{vanGinneken:2003uo}, a user directly manipulates points of the PDM, which has the disadvantage that the user needs to estimate the (unknown) corresponding position of the considered model point in the image. This is particularly difficult if the object does not exhibit distinct features. In their slice-wise SSM-based segmentation of abdominal aortic aneurysms, \citet{deBruijne:2004wm} initialise the current slice's PDM fitting from the segmentation result of the previous slice, with the option of manually correcting segmentations on a per-slice basis. \citet{Liu:2009gr} present Oriented Active Shape Models where the semi-automatic live-wire technique is coupled with an Active Shape Model. Other authors present tools for the posterior correction of model-based segmentations, such as \citet{Timinger:2003vf,Schwarz:2008hj,Tan:2014uo}. 
\citet{vandenHengel:2007ko} present an interactive procedure based on a set of rules and 2D sketches for completing partial 3D models in structure from motion.

To summarise, existing approaches performing surface reconstruction using SSMs either assume known correspondences or estimate point-based correspondences in an alternating manner. One exception is the work by \citet{Taylor:2016aa}, where, for sufficiently dense data, a surface-based hand pose model fitting is performed.

\section{Background}\label{background}
In this section we introduce the notation and define PDMs. %

\subsection{Notation}
$\onevec_p$ and $\zerovec_p$ denote the $p$-dimensional column vectors containing ones and zeros, respectively. $\matI_p$ denotes the $p \times p$ identity matrix and $\diag(\vecx)$ is the $p \times p$ matrix with the elements of the vector $\vecx \in \R^p$ on its diagonal. For a matrix $\matA$, $\matA_{i,j}$ is the (scalar) element in row $i$ and column $j$. The colon-notation is used to denote all rows or columns, e.g. $\matA_{:,j}$ is the $j$-th column of $\matA$. For a matrix $\matA$, the concatenation of all columns is given by $\VEC(\matA)$.

By $\prob(x)$ we denote the \emph{probability density function} (p.d.f.), or \emph{probability mass function} in the discrete case, where $x$ can indicate both a random variable and a realisation of it, depending on the context. $\mathcal{N}(x \vert \mu, \Sigma)$ denotes the Gaussian p.d.f. with mean $\mu$ and covariance matrix $\Sigma$.

\subsection{Point Distribution Models}\label{pdm}
Statistical shape models based on PDMs \citep{Cootes:1992uw} are an established technique to capture nonlinear shape deformations of shapes in $\R^3$ from training data by using a linear model in the higher-dimensional shape space. Let ${\{\matX_k : 1 \leq k \leq K\}}$ be the set of $K$ training shapes, where each shape is represented by $N$ points (or vertices) in $3$ dimensions given by the rows of the matrix $\matX_k \in \R^{N \times 3}$. 
For processing multiple shapes in a meaningful way, the $N$ vertices of all $K$ shapes have to be in correspondence, i.e.~the rows for all $\matX_k$ are corresponding to each other. 

The PDM is obtained by finding an (affine) subspace close to the subspace spanned by the training shapes, commonly performed by Principal Components Analysis (PCA).
For that, we define $\matX = [\vecx_1, \ldots, \vecx_K] \in \R^{3N \times K}$, with $\vecx_k = \VEC(\matX_k) \in \R^{3N}$.
This allows to compute the modes of shape variability as the eigenvectors of the sample covariance matrix $\matC = \frac{1}{K-1}(\matX - \bar{\vecx} \onevec_K^T) (\matX - \bar{\vecx} \onevec_K^T)^T$, where $\bar{\vecx} = \frac{1}{K} \sum_{k=1}^K \vecx_k$ is the mean of all shapes in the $3N$-dimensional shape space. Let $\matPhi \in \R^{3N \times M}$ be the matrix of the first $M$ eigenvectors of $\matC$ with largest eigenvalues. For $\vecalpha$ being a variable in $\R^M$, the PDM $\vecy(\vecalpha): \R^M \rightarrow \R^{3N}$ is a vector-valued function defined as
\begin{align} %
  \vecy(\vecalpha) = \bar{\vecx} + \matPhi \vecalpha \,,
\end{align}
where $\vecalpha \in \R^M$ is the shape deformation parameter. The deformation of vertex $i$ through $\vecalpha$ is denoted by
\begin{align}
  y_i(\vecalpha) = \bar{x}_i + \matPhi_i \vecalpha\,,
\end{align}
where the three rows of vertex $i$ are selected from $\bar{\vecx}$ and $\matPhi$ to obtain $\bar{x}_i \in \R^3$ and $\matPhi_i \in \R^{3 \times M}$.

A common assumption is that $\vecalpha$ follows a zero-mean Gaussian distribution,  %
i.e.~$\prob(\vecalpha) = \mathcal{N}(\vecalpha \vert \zerovec_M, \Lambda)$, where $\Lambda =  \diag(\lambda_1,\ldots, \lambda_M)$ with $\lambda_m$ being the $m$-th eigenvalue of $\matC$. Thus, thanks to imposing a distribution upon $\vecalpha$, we obtain a  distribution over shapes \citep{Albrecht:2013tw}.

Usually, in addition to the PDM accounting for shape deformations, a rigid transformation is employed in order to account for the absolute pose of the shape $\vecy(\vecalpha)$ with respect to some reference (e.g.~the image coordinate system). Here we assume that the predominant part of the pose has already been normalised. Minor pose variations can be (approximately) captured implicitly in the PDM.

\section{Problem Formulation}\label{probspec}
Given is a PDM that serves as prior for plausible shapes. Also, we assume that (for fixed $\vecalpha$) the PDM points are vertices of an oriented triangular surface mesh $\mathcal{M}$.
Additionally, we are given the set $\mathcal{P} = \{p_j: 1 \leq j \leq P\}$ of $P$ surface points of the shape that is to be reconstructed, where $\mathcal{P}$ is sparse in the sense that it only contains few points $p_j$ lying on the object's surface.
The objective is to find the deformation parameter $\vecalpha$ such that $\vecy(\vecalpha)$ ``best agrees'' with the points in $\mathcal{P}$.

\subsection{A Generative Model}
Our work is
motivated by the widely used Coherent Point Drift (CPD) approach for general point-set registration \citep{Myronenko:ve,Myronenko:2010wn}. It also resembles the approach by \citet{Zheng:2013ue} for deformable shape registration, which however is based on heteroscedastic GMMs with isotropic covariances, and it is related to the EM-ICP algorithm \citep{Granger:2002wb} since we also make use of hidden variables to model the unknown correspondences.

\begin{figure}[!h!t]
     \centerline{\includegraphics[scale=0.7]{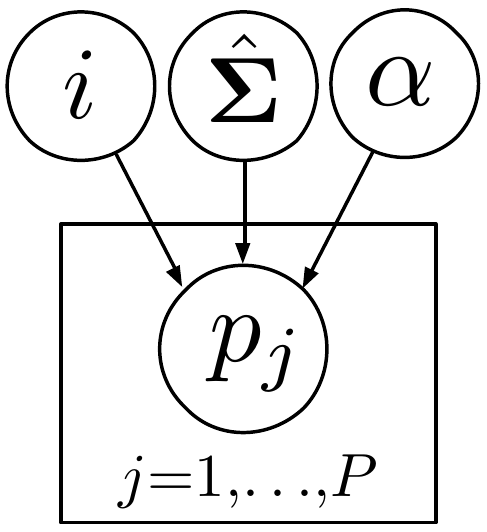}}
     \caption{Graphical model of the generating process of $\mathcal{P}$.}
     \label{graphicalmodel}
\end{figure}

In the following we present the probabilistic model, where a distribution is imposed over each vertex $i$ by using a GMM.
Given the set $\mathcal{P}$ and a PDM, we consider the following assumptions:
\begin{enumerate}
  \item For $i=1,\ldots,N$, each PDM vertex position $y_i(\vecalpha) \in \R^3$ is considered as the mean of a 3D Gaussian distribution with covariance $\matSigma_i$.
  \item Each point $p_j \in \mathcal{P}$ can be uniquely mapped to one specific vertex index $i$, its \emph{generating component}, from whose distribution it is drawn.
\end{enumerate}
As such, each point $p_j$ follows the distribution
\begin{align}\label{genmodel}
  \prob(p_j \vert i, \vecalpha, \matSigma_i) = \mathcal{N}(p_j \vert y_i(\vecalpha), \matSigma_i)\,,
\end{align}
where all $p_j$ for $j=1,\ldots,P$ are independent. The corresponding graphical model is depicted in Fig.~\ref{graphicalmodel}.
Note that all of the $N$ Gaussian components are parametrised by $\vecalpha$ and the covariances $\hat{\matSigma} := \{\matSigma_{i}\}$.
We assume that each component is chosen with equal probability, i.e. $\prob(i) = \frac{1}{N}$. 

Our objective is to find the parameters $\vecalpha$ and $\hat{\matSigma}$ that are most likely to have generated the points $\mathcal{P}$. Since the generating component $i_j$ of $p_j$ is unknown, the indices of the generating components are treated as \emph{latent} variables. 
To incorporate this uncertainty, we consider a GMM for the distribution of $p_j$, leading to
\begin{align}
   \prob(p_j \vert \vecalpha, \hat{\matSigma} )  & =  \sum_{i=1}^N \prob(i) ~ \prob(p_j \vert i, \vecalpha, \matSigma_i) \\
    & = \frac{1}{N} \sum_{i=1}^N \mathcal{N}(p_j \vert y_i(\vecalpha), \matSigma_i)\, . \label{objGen}
\end{align}
Using Bayes' theorem, one can derive the probability that the $i$-th mixture component has generated the point $p_j$, given also $\vecalpha$ and $\hat{\matSigma} $, as
\begin{align}
    \prob(i \vert p_j, \vecalpha, \hat{\matSigma} ) &= \frac{\prob(p_j \vert i, \vecalpha, \hat{\matSigma}) \prob(i) \prob(\vecalpha,\hat{\matSigma}) }{\sum_{i'}\prob(p_j \vert i', \vecalpha, \hat{\matSigma}) \prob(i')  \prob(\vecalpha,\hat{\matSigma})} \\
    & = \frac{\prob(p_j \vert i, \vecalpha, \matSigma_i) \prob(i)}{\sum_{i'}\prob(p_j \vert i', \vecalpha, \matSigma_{i'}) \prob(i') }\\
    &= \frac{\mathcal{N}(p_j \vert y_i(\vecalpha), \matSigma_i)}{\sum_{i'}\mathcal{N}(p_j \vert y_{i'}(\vecalpha), \matSigma_{i'})} \,. \label{mixweight}
\end{align}

\subsection{Optimisation using EM}
If the generating component $i_j$ of $p_j$ is unknown, according to eq.~\eqref{objGen} all points $p_j \in \mathcal{P}$ are independent and identically distributed (i.i.d.). Thus, the log-likelihood as a function of the model parameters $\vectheta := (\vecalpha, \hat{\matSigma})$ reads
\begin{align}
  L(\vectheta) = \ln \prob(\mathcal{P} \vert \vectheta ) &= \ln \prod_{j=1}^P \sum_{i=1}^N \prob(i) \prob(p_j \vert i, \vectheta )  \label{loglikelihood} \\
  &= \sum_{j=1}^P \ln  \sum_{i=1}^N \prob(i) \prob(p_j \vert i, \vectheta )  \,. \nonumber
\end{align}

The maximisation of $L$ w.r.t. $\vectheta$ cannot be solved readily due to the sum appearing inside the logarithm. Therefore, a common approach is to employ an iterative method for the maximisation. We denote the estimate of $\vectheta$ at iteration $n$ as $\vectheta^{(n)}$ and rewrite eq.~\eqref{loglikelihood} as
\begin{align}
  L(\vectheta, \vectheta^{(n)}) = \sum_{j=1}^P \ln  \sum_{i=1}^N \prob(i \vert p_j, \vectheta^{(n)}) \frac{\prob(i) \prob(p_j \vert i, \vectheta)}{\prob(i \vert p_j, \vectheta^{(n)})}\,.
\end{align}
Applying Jensen's inequality \citep{Jensen:1906up} leads to
\begin{align}
	& L(\vectheta, \vectheta^{(n)}) \geq  \label{jenineq}\\ 
	& \qquad \sum_{j=1}^P   \sum_{i=1}^N \prob(i \vert p_j, \vectheta^{(n)}) \ln \frac{\prob(i) \prob(p_j \vert i, \vectheta)}{\prob(i \vert p_j, \vectheta^{(n)})} =: Q(\vectheta, \vectheta^{(n)})\,. \nonumber 
\end{align}
As such, the right-hand side of eq.~\eqref{jenineq}, which we denote by $Q(\vectheta, \vectheta^{(n)})$, is a lower bound for $L(\vectheta, \vectheta^{(n)})$. Maximising this lower bound is the key idea of the EM algorithm. 
In the E-step, the probabilities $\prob(i \vert p_j,\vectheta^{(n)})$ are evaluated for fixed $\vectheta^{(n)}$ by using eq.~\eqref{mixweight}.
Then, in the M-step, $Q$ in eq.~\eqref{jenineq} is maximised w.r.t. $\vectheta$ for the fixed $\prob(i \vert p_j,\vectheta^{(n)})$ computed before.

\section{Methods}\label{methods} 
In section~\ref{anisoSurfaceReconstruction} below, we present the main novelty of this paper, the anisotropic GMM-based fitting approach that employs covariance matrices that ``oriented'' according to the surface normals at the PDM points. This allows to move from a purely point-based matching \citep{Myronenko:ve,Myronenko:2010wn,Zheng:2013ue,Bernard:2015wx} to a more surface-based fitting. Subsequently, we describe a fast approximation of the anisotropic GMM method. By using an extension to this approximation, one can ensure that it is an instance of the Generalised Expectation Maximisation (GEM) method and thus the convergence is guaranteed.

\subsection{Surface Reconstruction using an Anisotropic GMM}\label{anisoSurfaceReconstruction}
\begin{figure*}[!t]
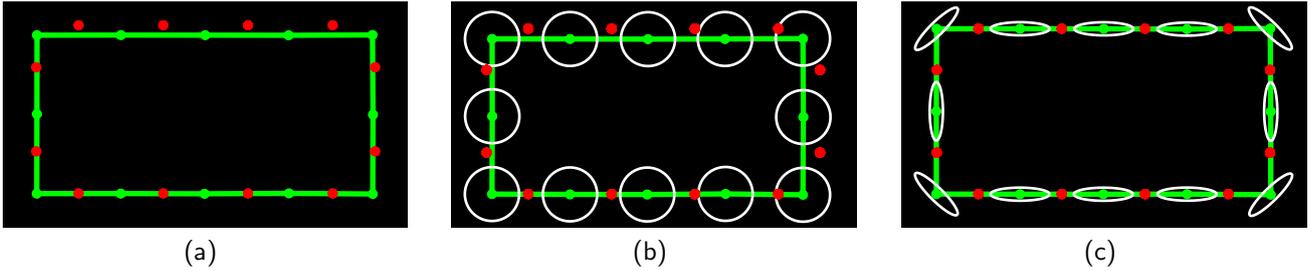
%
     \centerline{  
     \subfigure[]{\includetikz{didacticExample_init}} \hfil
     \subfigure[]{\includetikz{didacticExample_eta1}} \hfil 
     \subfigure[]{\includetikz{didacticExample_eta20}}  }
     \caption{Anisotropic covariance matrices to achieve a surface-based fitting. The sparse points $\mathcal{P}$ are shown in red, the PDM of a rectangle is shown in green, where the green points define the PDM vertices ($N=12, M=2, P=12)$. The orientation of the covariance matrices is shown as white ellipses. The objective is to deform the rectangle PDM such that it fits the red points by adjusting $\vecalpha$. The initialisation is shown in (a). Since the red points are sampled between the PDM points (cf.~\emph{shape approximation problem} \citep{hill95activeshape}), using spherical covariance matrices results in a fit that is even worse than the initialisation (b), whereas using anisotropic covariances results in a more accurate fit than the initialisation (c).}
     \label{didex}
\end{figure*}

When using spherical covariances for each of the $N$ Gaussian components, a purely point-based fitting is conducted. However, in a vast amount of medical applications of SSMs, the points of the PDM represent the vertices of a \emph{surface mesh}. This surface mesh is in general only an approximation of a \emph{continuous surface}. Whilst the sparse points $\mathcal{P}$ are assumed to lie on this continuous surface, in general they do not coincide with the PDM vertices. Hence, matching the \emph{surface}, depending on the PDM deformation parameter $\vecalpha$, is more appropriate. A didactic 2D example is presented in Fig.~\ref{didex}.

\subsubsection{Surface-aligned Covariance Matrices}
We now formalise our surface-based fitting method using a GMM with anisotropic covariance matrices. %
In the GMM, the covariance matrix of each component $i$, i.e. each vertex of the PDM, is defined as 
\begin{align}
	\matSigma_i(\sigma, \vecalpha) := \sigma^2 \matC_i(\vecalpha)\,.
\end{align}
 The scalar parameter $\sigma^2$ can be seen as a global scaling factor, whereas the matrix $\matC_i(\vecalpha)$ models the anisotropy of the surface structure locally by using a larger variance in the directions of vectors lying in the tangent plane of the PDM surface, compared to the variance along the PDM normal direction (cf.~Fig.~\ref{didex} (c)).

Assuming that the surface mesh $\mathcal{M}$ of the underlying shape of the PDM is given in the form of oriented triangles (cf.~section~\ref{probspec}), with $i_2$ and $i_3$ we denote the index of the ``left'' and ``right'' neighbour vertex of $i$, respectively. With that, the surface normal at vertex $i$ is given by
\begin{align}\label{normals}
  n_i(\vecalpha) = \frac{  (y_{i_2}(\vecalpha) - y_i(\vecalpha)) \times (y_{i_3}(\vecalpha) - y_i(\vecalpha)) }{\| (y_{i_2}(\vecalpha) - y_i(\vecalpha)) \times (y_{i_3}(\vecalpha) - y_i(\vecalpha)) \|} \,.
\end{align}
The matrix $\matC_i$ is defined as
\begin{align}\label{anisocov}
  \matC_i(\vecalpha) := (\frac{1}{\eta}-1)  n_i(\vecalpha)n_i^T(\vecalpha) + \matI_3 \,,
\end{align}
where the parameter $\eta \geq 1$ weights the variances of vectors along the normal direction compared to the tangential direction (note that we use the same value of $\eta$ for all points). A motivation for eq.~\eqref{anisocov} can be found in the work of \citet{hill95activeshape}.
For the covariance matrix $\matSigma_i(\sigma,\vecalpha)$, the variance along the normal $n_i(\vecalpha)$ is given by $\frac{\sigma^2}{\eta}$, and the variance in the direction of any vector in the tangent plane is $\sigma^2$. As such, for $\eta = 1$ one obtains an isotropic GMM fitting, in analogy to the CPD algorithm \citep{Myronenko:ve,Myronenko:2010wn}. We refer to the isotropic method with $\eta$ set to $1$ as \emph{ISO}. The two main differences between CPD and our approach are that in our case the transformations are parametrised by a PDM, and that we also allow for anisotropic covariances. Using transformations that are parametrised by a PDM has also been done by \citet{Zheng:2013ue} for deformable shape registration. 
Choosing $\eta > 1$ achieves the desired behaviour of modelling a larger variance in the tangent plane. Note that for $\eta>0$, the matrix $\matC_i(\vecalpha)$ is symmetric and positive definite with the spectrum $\{\frac{1}{\eta},1,1\}$. The inverse of $\matC_i$, the \emph{precision matrix}, is
\begin{align}\label{anisoW}
  \matW_i(\vecalpha) := \matC_i^{-1}(\vecalpha) = (\eta-1)  n_i(\vecalpha)n_i^T(\vecalpha) + \matI_3 \,.
\end{align}

\subsubsection{Maximum A Posteriori (MAP) Solution}
We can cast the log-likelihood from eq.~\eqref{loglikelihood} into a Bayesian view, leading to
\begin{align}
  L^{\text{posterior}}(\vectheta) = \ln [\prob(\mathcal{P} \vert \vectheta) \prob(\vectheta)] \,,
\end{align}
where additional knowledge in the form of the prior distribution $\prob(\vectheta)$ is incorporated. 
By assuming $\prob(\vecalpha) = \mathcal{N}(\vecalpha \vert \zerovec_M, \Lambda)$ as described in section~\ref{pdm}, and choosing a uniform prior for $\prob(\sigma)$, the prior distribution results in $\prob(\vectheta) \propto \prob(\vecalpha)$. The $Q$-function for the MAP solution reads
\begin{align}
   & Q(\vecalpha,\sigma,\vecalpha^{(n)},\sigma^{(n)}) = \label{qaniso}\\
  & \quad\quad\quad \text{const} - \frac{1}{2}\vecalpha^T \Lambda^{-1}\vecalpha - \frac{3P}{2}\ln\sigma^2 \nonumber\\
  & \quad\quad\quad\qquad -\frac{1}{2\sigma^2} \sum_{i,j} \prob(i \vert p_j, \vecalpha^{(n)}, \sigma^{(n)}) \cdot \nonumber\\
  & \qquad\quad\quad\quad\quad\quad\quad (p_j - y_i(\vecalpha))^T \matW_i(\vecalpha)(p_j - y_i(\vecalpha)) \nonumber\,.
\end{align}
As already described, the E-step is solved by evaluating eq.~\eqref{mixweight}. Then, the M-step comprises maximising $Q$ in eq.~\eqref{qaniso} w.r.t. $\vecalpha$ and $\sigma$. Due to the dependence of $\matW_i$ on $\vecalpha$ (for $\eta \neq 1$), finding $\vecalpha$ that maximises $Q$ does not admit a simple closed-form solution. This is in contrast to the isotropic case ($\eta = 1$), where $\vecalpha$ that maximises $Q$ can be found by solving a linear system of equations. Instead, $\vecalpha$ is now obtained using the BFGS quasi-Newton method \citep{Nocedal:2006uv}. The idea is to start with the old value $\vecalpha^{(n)}$, and then iteratively move along directions that increase $Q$. 
Whilst the ordinary Newton method requires the gradient of $Q$ as well as its Hessian, the BFGS quasi-Newton method uses an approximation of the Hessian that is cheap to compute. 
A derivation of the gradient $\nabla_{\vecalpha} Q$ of $Q$ w.r.t. $\vecalpha$ can be found in~\ref{appendixGradient}.

In order to obtain $\vecalpha$, one option is to run the BFGS quasi-Newton procedure until convergence, where one obtains an $\vecalpha$ that (locally) maximises $Q$. With that, the updates of $\sigma$ on $\vecalpha$ depend on each other and the procedure reverts to the ECM algorithm \citep{Meng:1993uq}. %
An alternative is to run only a single quasi-Newton step in each M-step. With that, the obtained $\vecalpha$ is not a local maximiser of $Q$; however, one still has the guarantee that $Q$ is non-decreasing. As such, this procedure reverts to the GEM algorithm \citep{Dempster:1977ul}.

Finally, the $\sigma$-update for fixed $\vecalpha$ is given by
\begin{align}
  \sigma^2 = & \frac{1}{3P} \sum_{i,j} \prob(i \vert p_j, \vecalpha^{(n)}, \sigma^{(n)}) \cdot \\
  & \quad\quad\quad\quad (p_j - y_i(\vecalpha))^T \matW_i(\vecalpha)(p_j - y_i(\vecalpha))\nonumber \,.
\end{align}  
The pseudocode of the anisotropic GMM fitting procedure is presented in Algorithm~\ref{anisoalg}.

\begin{algorithm}[h!]\label{anisoalg}
\scriptsize
\SetKwInput{Input}{Input}
\SetKwInput{Output}{Output}
\SetKwInput{Initialise}{Initialise}
\DontPrintSemicolon

 \Input{$\bar{\vecx}, \matPhi, \mathcal{P}, \eta, \mathcal{M}$}
 \Output{$\vecalpha, \sigma^2$}
\Initialise{$\vecalpha  = \zerovec$, $\sigma^2 = \frac{1}{3NP} \sum_{i,j} \|p_j - \bar{x}_i\|^2$, $\matP \in \R^{P \times N}$, $\vecy = \bar{\vecx} + \matPhi \vecalpha$}
\ForEach{$i=1,\ldots,N$}{  
   \tcp{compute $\matW_i$}
   $n_i= \frac{  (y_{i_2} - y_i) \times (y_{i_3} - y_i) }{\| (y_{i_2} - y_i) \times (y_{i_3} - y_i) \|}$\\
   $\matW_{i} = (\eta-1)  n_in_i^T + \matI_3 $ %
   }
 \Repeat{convergence}{
   \tcp{E-step}
  \ForEach{$j=1,\ldots,P$}{
   $t = 0$\\
   \ForEach{$i=1,\ldots,N$}{
   $\matP_{ji}  = \exp(-\frac{1}{2 \sigma^2} (p_j - y_i)^T \matW_i (p_j - y_i))$ \\
   $t = t + \matP_{ji}$
   }
   $\matP_{j,:} = \frac{1}{t}\matP_{j,:}$
   }
   \BlankLine
   \tcp{M-step}
   $\vecalpha = $ quasi-Newton($Q, \nabla Q, \vecalpha$) \label{quasinewtonupdate}\\
   $\vecy = \bar{\vecx} + \matPhi \vecalpha$\\
    \ForEach{$i=1,\ldots,N$}{  
       \tcp{compute $\matW_i$}
       $n_i= \frac{  (y_{i_2} - y_i) \times (y_{i_3} - y_i) }{\| (y_{i_2} - y_i) \times (y_{i_3} - y_i) \|}$\\
       $\matW_{i} = (\eta-1)  n_in_i^T + \matI_3 $ %
       } 

   $\sigma^2 = \frac{1}{3P} \sum_{i,j} \matP_{ji} (p_j - y_i)^T \matW_i(p_j - y_i)$
   }
 \caption{Pseudocode of the anisotropic GMM fitting method. The notation ``quasi-Newton($Q, \nabla Q, \vecalpha$)'' denotes running the quasi-Newton method for maximising $Q$ w.r.t. $\vecalpha$, where $\nabla Q$ is its gradient and the third argument is the initial value of $\vecalpha$. If the GEM approach is used, the quasi-Newton method is run only for a single iteration. Note that the surface mesh $\mathcal{M}$ is used for the normal computations.}
\end{algorithm}

\subsubsection{Fast Approximate Anisotropic GMM}
Now we introduce an approximation of the $\vecalpha$-update that is a much faster alternative to the quasi-Newton method. 
The main idea is to use the previous value  $\vecalpha^{(n)}$ instead of $\vecalpha$ for computing the anisotropic covariance matrices $\matC_i(\vecalpha^{(n)})$ during the $\vecalpha$-update in the M-step. Our key assumption is that the PDM is \emph{well-behaved} in the sense that neighbouring vertices vary smoothly during deformation; thus, \emph{locally} the deformation of an individual triangle is nearly a translation. Since surface normals are \emph{invariant} to translations it follows that ${\|n_{i}(\vecalpha)-n_{i}(\vecalpha^{(n)})\|}$ is small, which implies that ${\|W_i(\vecalpha)-W_i(\vecalpha^{(n)})\|}$ is also small.

The resulting $Q$-function using the proposed approximation is now given by
\begin{align}
  & \tilde Q(\vecalpha,\sigma,\vecalpha^{(n)},\sigma^{(n)}) = \label{qanisoapprox} \\
  & \quad\quad\quad \text{const} - \frac{1}{2}\vecalpha^T \Lambda^{-1}\vecalpha - \frac{3P}{2}\ln\sigma^2 \nonumber \\
  & \quad\quad\quad\qquad -\frac{1}{2\sigma^2} \sum_{i,j} \prob(i \vert p_j, \vecalpha^{(n)}, \sigma^{(n)}) \cdot \nonumber\\
  & \qquad\quad\quad\quad\quad\quad\quad (p_j - y_i(\vecalpha))^T \matW_i(\vecalpha^{(n)})(p_j - y_i(\vecalpha)) \nonumber\,,
\end{align}
where the difference to $Q$ in eq.~\eqref{qaniso} is that the constant $\matW_i(\vecalpha^{(n)})$ is now used in place of the function $\matW_i(\vecalpha)$. As such, the $\vecalpha$-update in the M-step is a quadratic concave problem that can be maximised efficiently. The solution for $\vecalpha$ is found by solving the linear system $ \matA \vecalpha =  \vecb$, where $ \matA \in \R^{M \times M}$ is given by
\begin{align}
	& \matA = \sigma^2\Lambda^{-1} + \sum_{i,j} \prob(i \vert p_j, \vecalpha^{(n)}, \sigma^{(n)}) \matPhi_{i}^T \matW_i(\vecalpha^{(n)})\matPhi_{i} \,,
	\intertext{and $ \vecb \in \R^M$ by}
	& \vecb = \sum_{i,j} \prob(i \vert p_j, \vecalpha^{(n)}, \sigma^{(n)})\matPhi_{i}^T \matW_i(\vecalpha^{(n)}) (p_j - \bar{x}_i)  \,.
\end{align}
The pseudocode for this approximate method is similar to Algorithm~\ref{anisoalg}, except for line~\ref{quasinewtonupdate}, where $\vecalpha$ is computed by solving a linear system.

In order to guarantee that the approximate method converges, it is necessary that in the M-step the value of the \emph{exact} $Q$ in eq.~\eqref{qaniso} is non-decreasing, i.e. the new $\vecalpha =  \matA^{-1}  \vecb$ obtained using $\tilde Q$ must fulfil
\begin{align}\label{approxconvcond}
  Q( \matA^{-1}  \vecb,\sigma^{(n)}, \vecalpha^{(n)}, \sigma^{(n)}) \geq Q(\vecalpha^{(n)},\sigma^{(n)}, \vecalpha^{(n)}, \sigma^{(n)}) \,.
\end{align}
For $\eta = 1$ the methods reverts to the isotropic case and condition~\eqref{approxconvcond} vacuously holds. However, for $\eta>1$ this is not true in general. One way to ensure that $Q$ is non-decreasing is to evaluate the condition in eq.~\eqref{approxconvcond} in each iteration, and, in the case of a violation, revert to one of the quasi-Newton methods for the $\vecalpha$-update. Specifically, we consider a single quasi-Newton step for updating $\vecalpha$. We denote the approximate method without this convergence check as \emph{ANISO}, and the method with the convergence check and the quasi-Newton step as fall-back as \emph{ANISOc}.

\begin{figure*}[!ht]
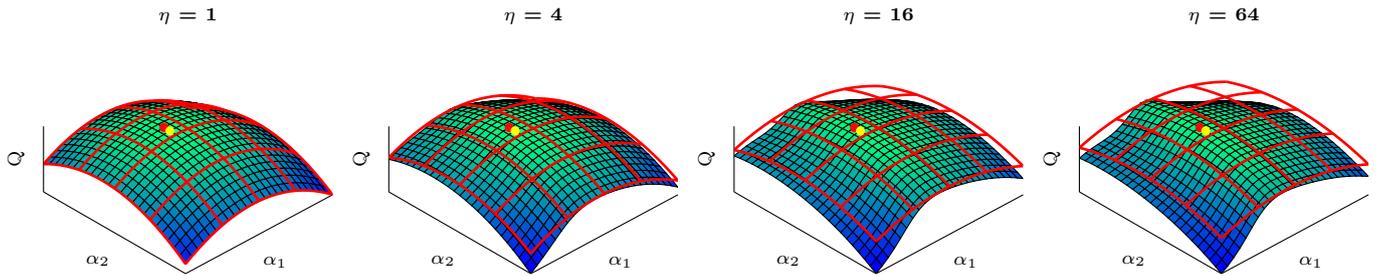

    \centerline{  
       \subfigure{\includetikz{nonconcavity_plots-eta_1}} \hfil
       \subfigure{\includetikz{nonconcavity_plots-eta_4}} \hfil
       \subfigure{\includetikz{nonconcavity_plots-eta_16}} \hfil
       \subfigure{\includetikz{nonconcavity_plots-eta_64}}
     }
     \caption{Illustration of the behaviour of $Q$ for various $\eta$. The height and the colour of the surface both show the value of $Q$, eq.~\eqref{qaniso}, depending on $\vecalpha_1$ and $\vecalpha_2$. The red grid shows its concave approximation $\tilde{Q}$ as presented in eq.~\eqref{qanisoapprox}. The red dot denotes the value of $Q$ at $\vecalpha^{(n)}$; at this position $Q = \tilde{Q}$. The yellow dot indicates the maximum of $Q$. For the trivial case of $\eta =1$ it can be seen that $Q = \tilde{Q}$ everywhere, whereas an increasing $\eta$ leads to a larger discrepancy between $Q$ and $\tilde{Q}$ as well as to an ``increased non-concavity'' of $Q$.}
     \label{nonconcavity}
\end{figure*}

\begin{table*}[!hb]
\scriptsize
\caption{Computational complexity table. The complexity of the $\vecalpha$-update for one iteration of the BFGS quasi-Newton methods is $\mathcal{O}(M^2)$ plus the complexity of the evaluation of $Q$ and $\nabla Q$ \citep{Nocedal:2006uv} (we use $n$ to denote the number of iterations of the quasi-Newton method). The complexity of the $\vecalpha$-update of the remaining methods comprises the computation of $\matA$ and $\vecb$, as well as solving a linear system of equations of size $M\times M$, for which we present the complexity $\mathcal{O}(M^3)$ due to the matrix inversion involved. Note that in $^{\star}$ we present the complexity for general $\Lambda$, for diagonal $\Lambda$ the quadratic time complexity in $M$ reduces to linear complexity.}
\centerline{
\begin{tabular}{ll|c|c|ccc}
\toprule
                                    &                                          & \multicolumn{4}{c|}{anisotropic}                                                                                                 & isotropic            \\
                                    &                                          & \multicolumn{1}{l|}{ECM}   & \multicolumn{1}{l|}{GEM}   & \multicolumn{1}{l|}{ANISOc}   & \multicolumn{1}{l|}{ANISO} & \multicolumn{1}{c}{ISO} \\ \midrule
update $\vecy$                      &                                          & \multicolumn{5}{c}{$\mathcal{O}(MN)$}                                                                                                                            \\ \midrule
compute $\{\matW_i\}$               &                                          & \multicolumn{4}{c|}{$\mathcal{O}(N)$}                                                                                                      & -                    \\ \midrule
E-step                              &                                          & \multicolumn{5}{c}{$\mathcal{O}(NP)$}                                                                                                                            \\ \midrule
\multirow{5}{*}{$\vecalpha$-update} & evaluate $Q$, eq.~\eqref{qaniso}       & \multicolumn{3}{c|}{$\mathcal{O}(MN + M^2 + NP)^{\star}$ } & \multicolumn{2}{c}{-}                              \\ \cmidrule{2-7} 
                                    & evaluate $\nabla Q$, eq.~\eqref{gradq} & \multicolumn{3}{c|}{$\mathcal{O}(M^2 + MNP)^{\star}$ }               & \multicolumn{2}{c}{-}                              \\ \cmidrule{2-7} 
                                    & construct $\matA $                          & \multicolumn{2}{c|}{-}                                  & \multicolumn{3}{c}{$\mathcal{O}(M^2N + NP)$}                                                                          \\ \cmidrule{2-7} 
                                    & construct $\vecb $                         & \multicolumn{2}{c|}{-}                                   & \multicolumn{3}{c}{$\mathcal{O}(MN+NP)$}                                                                          \\ \cmidrule{2-7} 
                                    & total $\vecalpha$-update                 & $\mathcal{O}(n(M^2 + MNP))$          & $\mathcal{O}(M^2 + MNP)$             & $\mathcal{O}(M^3 + MNP +  M^2N)$ &\multicolumn{2}{|c}{$\mathcal{O}(M^3+M^2N+NP)$}                                                  \\ \midrule
$\sigma$-update                     &                                          & \multicolumn{5}{c}{$\mathcal{O}(NP)$}                                                                                                                            \\ \midrule
total &  (per outer iteration)                                        & $\mathcal{O}(n(M^2 + MNP))$          & $\mathcal{O}(M^2 + MNP)$             & \multicolumn{1}{c|}{$\mathcal{O}(M^3+MNP+M^2N)$} & \multicolumn{2}{c}{$\mathcal{O}(M^3+M^2N+NP)$}                                                         \\ \bottomrule
\label{convergenceTable}
\end{tabular}
}
\end{table*}

In Fig.~\ref{nonconcavity} we illustrate the behaviour of $Q$ and compare it with $\tilde{Q}$ for various choices of $\eta$. Note that for visualisation purposes we have chosen $M=2$, whereas in higher-dimensional cases the effect of an increasing $\eta$ on the non-concavity can be expected to be more severe.

\subsection{Performance Analysis}
Table~\ref{convergenceTable} summarises the computational complexity of the presented methods. 
In Fig.~\ref{convergencetime} we plot the mean of the normalised value of $Q$ as a function of the processing time for all four anisotropic fitting methods. For each random run we sample a shape instance by drawing $\vecalpha$ (cf.~section~\ref{pdm}), select $P$ points randomly from the mesh surface (cf.~section~\ref{drawRandomTrianglePts}), and run the four methods. The obtained $Q$ for the four methods are then normalised such that in each run the smallest $Q$ corresponds to $0$ and the largest $Q$ corresponds to $1$ (normalisation w.r.t to all four methods simultaneously).
We have found that the single-step quasi-Newton method (GEM) is faster compared to the full quasi-Newton procedure (ECM). Moreover, compared to both quasi-Newton methods, the approximate methods are much faster. Since the ANISOc method makes use of elements both of the GEM and the ANISO method, the total time complexity of the ANISOc method is the combined time complexity for GEM and ANISO (cf.~Table~\ref{convergenceTable}). Nevertheless, in our simulations the ANISOc method comes close to the ANISO method in terms of convergence speed. 
This is because in the early stages of the iterative procedure the ANISOc method satisfies condition \eqref{approxconvcond} in most cases. A violation of \eqref{approxconvcond} happens more frequently in the later stages. %
Since these results suggest that the ANISO method is faster and as good as the other methods, for our experiments we use the ANISO method as representative for the anisotropic fitting methods. %

 \newcommand{\convergenePlotScale}{.77}

\begin{figure*}
\centerline{\includetikz{convergencePlots_legend}} 
\centerline{
       \subfigure{\includegraphics[scale=\convergenePlotScale]{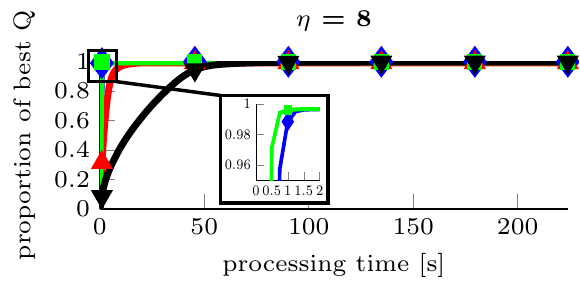}} \hfil
       \subfigure{\includegraphics[scale=\convergenePlotScale]{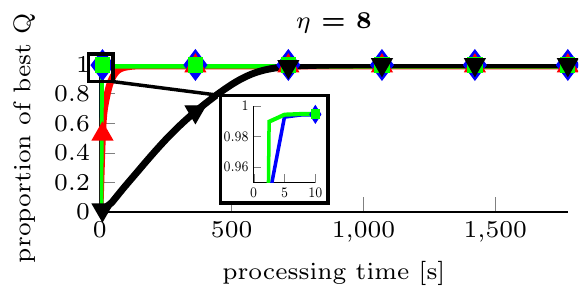}} \hfil
       \subfigure{\includegraphics[scale=\convergenePlotScale]{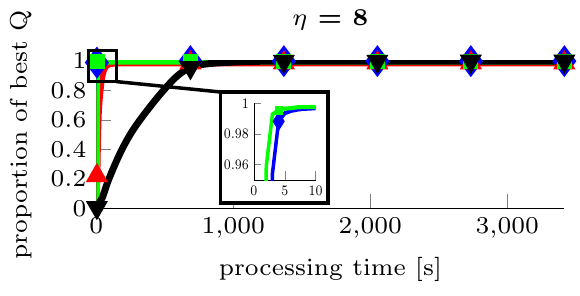}} \hfil
       \subfigure{\includegraphics[scale=\convergenePlotScale]{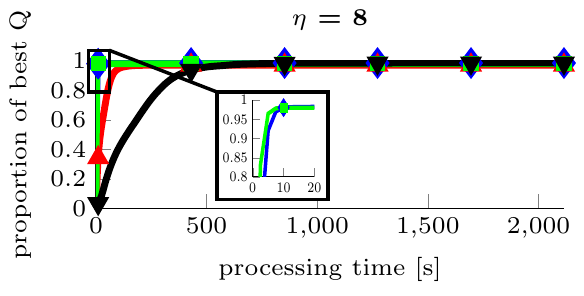}} 
     } 
     \centerline{
       \subfigure{\includegraphics[scale=\convergenePlotScale]{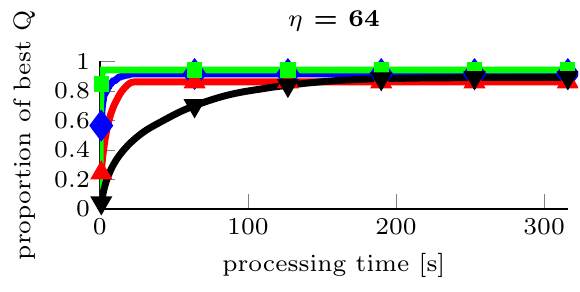}} \hfil
       \subfigure{\includegraphics[scale=\convergenePlotScale]{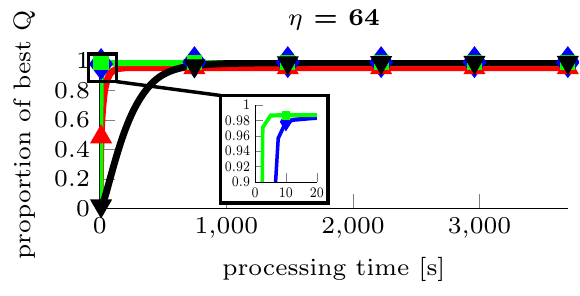}} \hfil
       \subfigure{\includegraphics[scale=\convergenePlotScale]{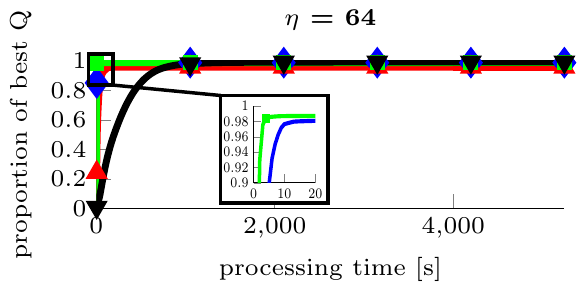}} \hfil
       \subfigure{\includegraphics[scale=\convergenePlotScale]{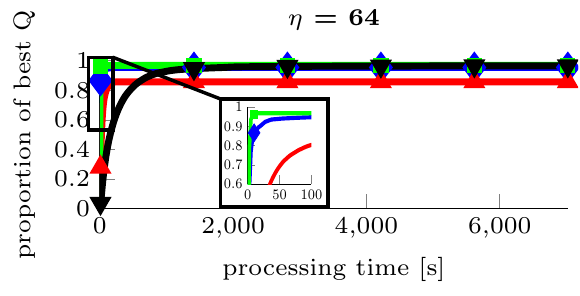}} 
     } 
 \caption{Proportion of best value of $Q$ versus processing time averaged over $100$ random runs for the four anisotropic methods  for two choices of $\eta$ (in the rows). In each column a different dataset has been used to produced the results, from left to right we show results produced by the brain shapes dataset ($N=1792,M=16,P = 30$, cf.~section~\ref{brainShapes}), the femur dataset ($N=3800,M=59,P = 30$, cf.~section~\ref{femur}), the tibia dataset ($N=4071,M=59,P = 30$, cf.~section~\ref{femur}), and the hip dataset ($N=5603,M=47,P = 30$, cf.~section~\ref{hip}).}
 \label{convergencetime}
\end{figure*}

\section{Experiments}\label{experiments}
\label{drawRandomTrianglePts}
In this section we evaluate the proposed fitting procedures on five datasets with the parameters being shown in Table~\ref{datasettable}. For the generation of the set of sparse points $\mathcal{P}$, we sample sparse points randomly on the shape \emph{surfaces}. 
To do so, we first select a triangle from the surface mesh with a probability proportional to its area. Then, we uniformly sample a point lying within the triangle according to the procedure presented by \citet{Osada:2002tq}. Moreover, in order to evaluate how well our method is able to cope with uncertainties in the given points, we considered noisy versions of these points by adding spherical Gaussian noise with covariance $\tilde{\sigma}^2 \matI_3$ to each point individually. For each experiment we have taken the scale of the object in the shape model into account for choosing $\tilde{\sigma}$. The considered values of $\tilde{\sigma}$ are presented in Table~\ref{datasettable}.

\begin{table}[h!]
\scriptsize
\centering
\caption{Summary of parameters for the datasets. $N$ is the number of points in the PDM (``ds'' denotes the downsampled PDM), $K$ the number of training shapes, $M$ the number of modes of variation for leave-all-in (LAI) and leave-one-out (LOO) experiments, $\eta$ the anisotropy parameter, and $\tilde{\sigma}$ the standard deviation of the noise added to the points.}
\label{datasettable}
\begin{tabular}{@{}lccccccc@{}}
\toprule
             & $N$    & $N$ (ds) & $K$  & $M$ (LAI) & $M$ (LOO) & $\eta$  & $\tilde{\sigma}$   \\ \midrule
brain  		 & $1792$ & $371$    & $17$ & $16$    & $96$ (kPCA) & $8$ & $2$~mm\\
femur        & $3800$ & $759$    & $60$ & $59$    & $58$        & $4$ & $5$~mm  \\
tibia        & $4701$ & $814$    & $60$ & $59$    & $58$   		& $4$ & $5$~mm \\
hip          & $5603$ & $1120$   & $48$ & $47$    & $46$  		& $4$ & $5$~mm \\
liver        & $4542$ & $908$    & $112$ & $111$  & $110$ 		& $2$ & $1$~cm\\ \bottomrule
\end{tabular}
\end{table}

In addition to the proposed probabilistic fitting methods presented in this paper, we also evaluate two non-probabilistic ICP approaches. The first approach is the regularised isotropic ICP method as outlined in Algorithm~{\ref{icpalg}, 
which we denote \emph{ICP} in the experiments. The second approach is an anisotropic version thereof, which we denote \emph{AICP} in the experiments. Similarly to \cite{MaierHein:2010uk}, 
for the anisotropic ICP we compute the nearest neighbours in the third line of Algorithm~{\ref{icpalg}} by taking the anisotropic covariance matrices (eq.~{\eqref{anisocov}}) into account.
\begin{algorithm}[t!]\label{icpalg}
\scriptsize

\SetKwInput{Input}{Input}
\SetKwInput{Output}{Output}
\SetKwInput{Initialise}{Initialise}
\DontPrintSemicolon

 \Input{$\bar{\vecx}, \matPhi, \mathcal{P}, \Lambda$}
 \Output{$\vecalpha$}
\Initialise{$\vecalpha  = \zerovec$, $\vecp = \VEC([p_1, \ldots,p_P]) \in \R^{3P}$}
 \Repeat{convergence}{
 $\vecy = \bar{\vecx} + \matPhi \vecalpha$ \\
   \tcp{Find nearest neighbours}
   $\mathcal{N} = $ findNearestNeighbourIndices($\vecy$, $\mathcal{P}$)\\
   \BlankLine
   \tcp{Solve linear system for $\vecalpha$}
   $\matA = \matPhi_{\mathcal{N},:}$\\
   $\vecb = \vecp - \bar{\vecx}_{\mathcal{N}}$\\
   $\vecalpha = (\matA^T \matA + \Lambda^{-1})^{-1}\matA^T \vecb$ \tcp*{Tikhonov regularisation}
   
   }
 \caption{Pseudocode of the ICP baseline method. The notation $\matPhi_{\mathcal{N},:}$ and $\bar{\vecx}_{\mathcal{N}}$ means selecting the appropriate rows from $\matPhi$ and $\bar{\vecx}$ according to the indices of the nearest neighbours $\mathcal{N}$.}
\end{algorithm}
For both, ICP and AICP, the regulariser $\Lambda$ corresponds to the covariance matrix of the PDM parameter $\vecalpha$ (cf.~section~\ref{pdm}).}
Moreover, in our evaluation we compare the ground truth data to the mean shape,~i.e. in this setting we do not run any fitting procedure at all, which amounts to setting $\vecalpha = \zerovec$.

The anisotropic method requires to set the parameter $\eta$ accounting for the amount of anisotropy. We have manually chosen the values of $\eta$ for each dataset, as shown in Table~\ref{datasettable}. We have empirically found that for an increasing amount of uncertainty $\tilde{\sigma}$ in the sparse points, it is advantageous to use a lower value of $\eta$.

We consider leave-all-in (LAI) and leave-one-out (LOO) experiments. The LAI experiments measure the performance of our method given a perfect model, whereas the LOO experiments evaluate the generalisation ability to unseen data.

We use the Dice Similarity Coefficient (DSC) as volumetric overlap measure, which is defined as
\begin{align}
	DSC(\mathcal{V}_x, \mathcal{V}_y) = \frac{2 \vert \mathcal{V}_x \cap \mathcal{V}_y \vert}{ \vert \mathcal{V}_x \vert +  \vert\mathcal{V}_y \vert} 
\end{align}
for the volumetric segmentations $\mathcal{V}_x$ and $\mathcal{V}_y$. Additional results considering surface-based measures for all datasets are presented in the supplementary material.

\ifRowPlots
\newcommand{\meanStdPlotsA}[3] {
\begin{figure}
\centerline{\subfigure{\includetikz{#1_dice_LAI_rnd_meanStdPlots_legend}}}
\centerline{
\hspace{2.5cm}
       \begin{minipage}{0.5\textwidth}
      \centerline{%
      \subfigure{\includetikz{#1_dice_LAI_rnd_mean}}%
      \subfigure{\includetikz{#1_runtime_LAI_rnd_mean}}%
      }
      \vspace{-3mm}
      \centerline{%
      \subfigure{\includetikz{#1_dice_LAI_rnd_std}}%
      \subfigure{\includetikz{#1_runtime_LAI_rnd_std}}%
      }
       \vspace{-4mm}
       \caption{DSC and runtime for #2 LAI results.}
       \label{#1ResultsLAISummary}
      \end{minipage}
\hfil
    \begin{minipage}{1\textwidth}
    \centerline{%
    \subfigure{\includetikz{#1_dice_LOO#3_rnd_mean}}%
    \subfigure{\includetikz{#1_runtime_LOO#3_rnd_mean}}%
    }
    \vspace{-3mm}
    \centerline{%
    \subfigure{\includetikz{#1_dice_LOO#3_rnd_std}}%
    \subfigure{\includetikz{#1_runtime_LOO#3_rnd_std}}%
    }\vspace{-4mm}
         \caption{DSC and runtime for #2 LOO results.}
         \label{#1ResultsLOOSummary}
    \end{minipage}
}
\end{figure} 
 }

\newcommand{\meanStdPlotsB}[3] {
\begin{figure}
\centerline{\subfigure{\includetikz{#1_dice_LAI_rnd_meanStdPlots_legend}}}
\centerline{
\hspace{2.5cm}
       \begin{minipage}{0.5\textwidth}
      \centerline{%
      \subfigure{\includetikz{#1_S-avg_LAI_rnd_mean}}%
      \subfigure{\includetikz{#1_S-max_LAI_rnd_mean}}%
      }
      \vspace{-3mm}
      \centerline{%
      \subfigure{\includetikz{#1_S-avg_LAI_rnd_std}}%
      \subfigure{\includetikz{#1_S-max_LAI_rnd_std}}%
      }
       \vspace{-4mm}
       \caption{Surface distances for #2 LAI results.}
       \label{#1ResultsLAISummaryB}
      \end{minipage}
\hfil
    \begin{minipage}{1\textwidth}
    \centerline{%
    \subfigure{\includetikz{#1_S-avg_LOO#3_rnd_mean}}%
    \subfigure{\includetikz{#1_S-max_LOO#3_rnd_mean}}%
    }
    \vspace{-3mm}
    \centerline{%
    \subfigure{\includetikz{#1_S-avg_LOO#3_rnd_std}}%
    \subfigure{\includetikz{#1_S-max_LOO#3_rnd_std}}%
    }\vspace{-4mm}
         \caption{Surface distances for #2 LOO results.}
         \label{#1ResultsLOOSummaryB}
    \end{minipage}
}
\end{figure} 

}
\else

\newcommand{\meanStdPlotsA}[4] {
       \begin{figure}[#4]
  \centerline{\subfigure{\includetikz{#1_dice_LAI_rnd_meanStdPlots_legend}}}
  \vspace{-3mm}
\centerline{\begin{sideways}{\qquad\qquad\qquad \scriptsize LAI}\end{sideways}%
\subfigure{\includetikz{#1_dice_LAI_rnd_mean}} \hfil%
\subfigure{\includetikz{#1_dice_LAI_rnd_std}}\hfil%
\subfigure{\includetikz{#1_runtime_LAI_rnd_mean}}%
}
\vspace{-2mm}
\centerline{\begin{sideways}{\qquad\qquad\qquad \scriptsize LOO}\end{sideways}%
\subfigure{\includetikz{#1_dice_LOO#3_rnd_mean}}\hfil%
\subfigure{\includetikz{#1_dice_LOO#3_rnd_std}}\hfil%
\subfigure{\includetikz{#1_runtime_LOO#3_rnd_mean}}%
}
\vspace{-3mm}
 \caption{DSC and runtime for #2 LAI and LOO results.}
 \label{#1ResultsSummary}

  \end{figure} 
}

\newcommand{\meanStdPlotsB}[4] {
\begin{figure*}[#4]
\centerline{\subfigure{\includetikz{#1_dice_LAI_rnd_meanStdPlots_legend}}}
\vspace{-3mm}
\centerline{\begin{sideways}{\qquad\qquad\qquad \scriptsize LAI}\end{sideways}%
\subfigure{\includetikz{#1_S-avg_LAI_rnd_mean}}\hfil%
\subfigure{\includetikz{#1_S-avg_LAI_rnd_std}}\hfil
\subfigure{\includetikz{#1_S-max_LAI_rnd_mean}}\hfil%
\subfigure{\includetikz{#1_S-max_LAI_rnd_std}}
}
\vspace{-2mm}
\centerline{\begin{sideways}{\qquad\qquad\qquad \scriptsize LOO}\end{sideways}%
\subfigure{\includetikz{#1_S-avg_LOO#3_rnd_mean}}\hfil%
\subfigure{\includetikz{#1_S-avg_LOO#3_rnd_std}}\hfil%
\subfigure{\includetikz{#1_S-max_LOO#3_rnd_mean}}\hfil%
\subfigure{\includetikz{#1_S-max_LOO#3_rnd_std}}
}
\vspace{-3mm}  
\caption{Surface distances for #2 LAI and LOO results.}
\label{#1ResultsSummaryB}
\end{figure*} 
}
\fi

\subsection{Knee Bones: Femur and Tibia}\label{femur}
\newcommand{\femurTibiaScaleA}{.18}
\newcommand{\tibiaScaleA}{.18}
\begin{figure}[!t!]  
     \centerline{    
       \subfigure[]{\includegraphics[scale=\femurTibiaScaleA]{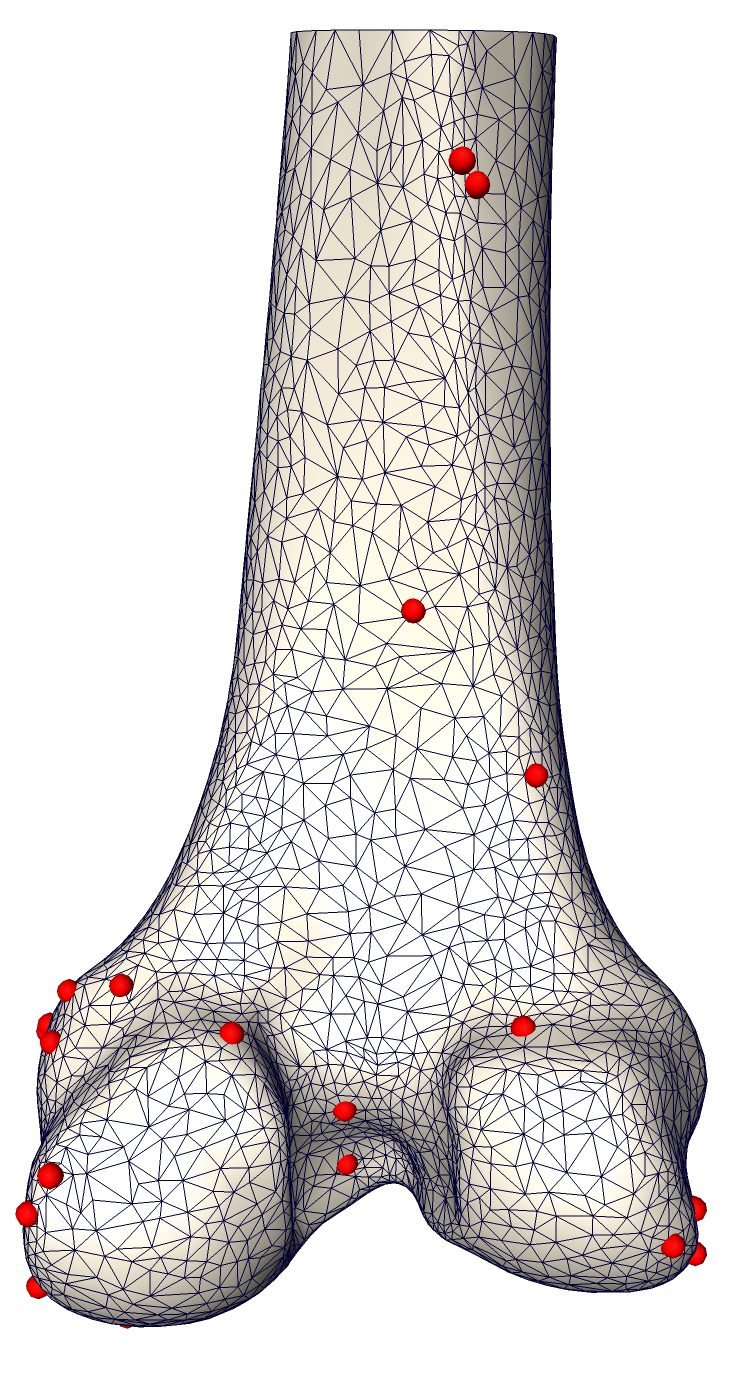}} \hfil  
       \subfigure[]{\includegraphics[scale=\femurTibiaScaleA]{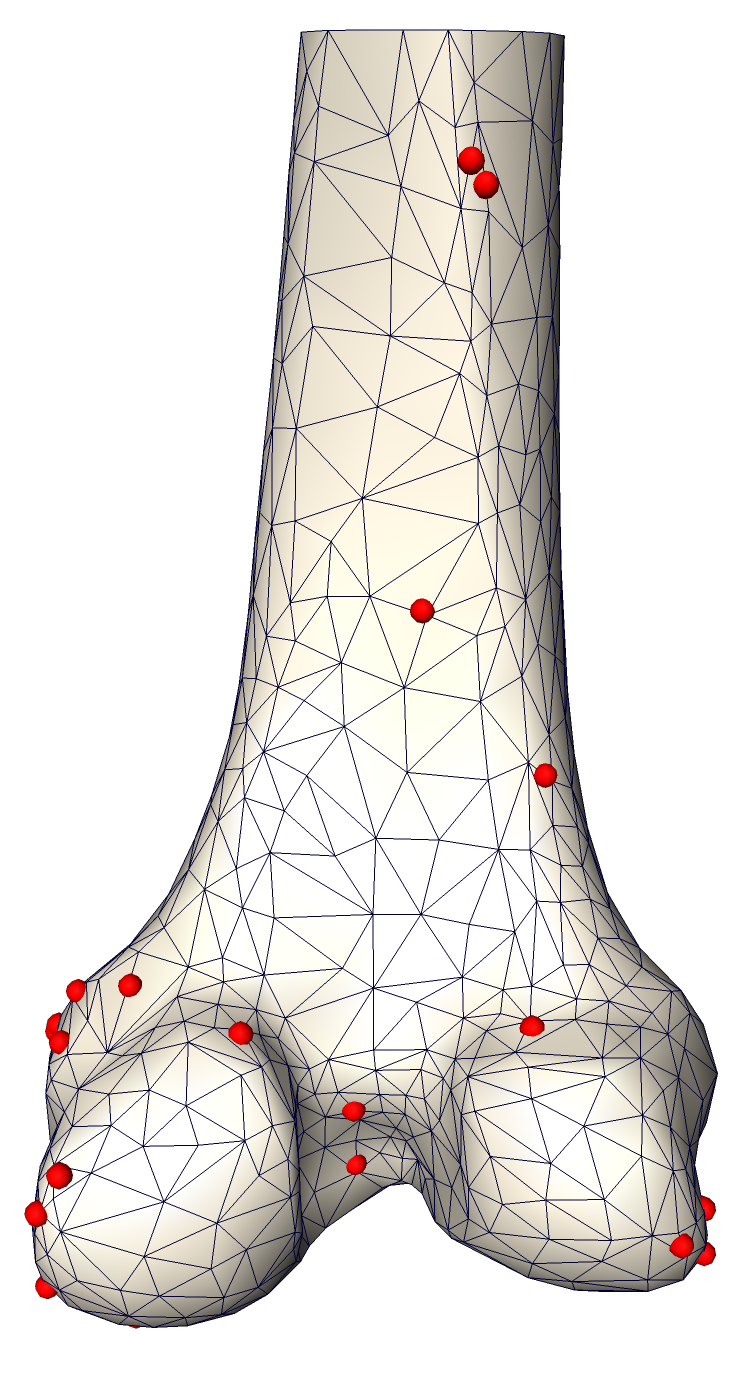}}
     }
     \centerline{    
       \subfigure[]{\includegraphics[scale=\tibiaScaleA]{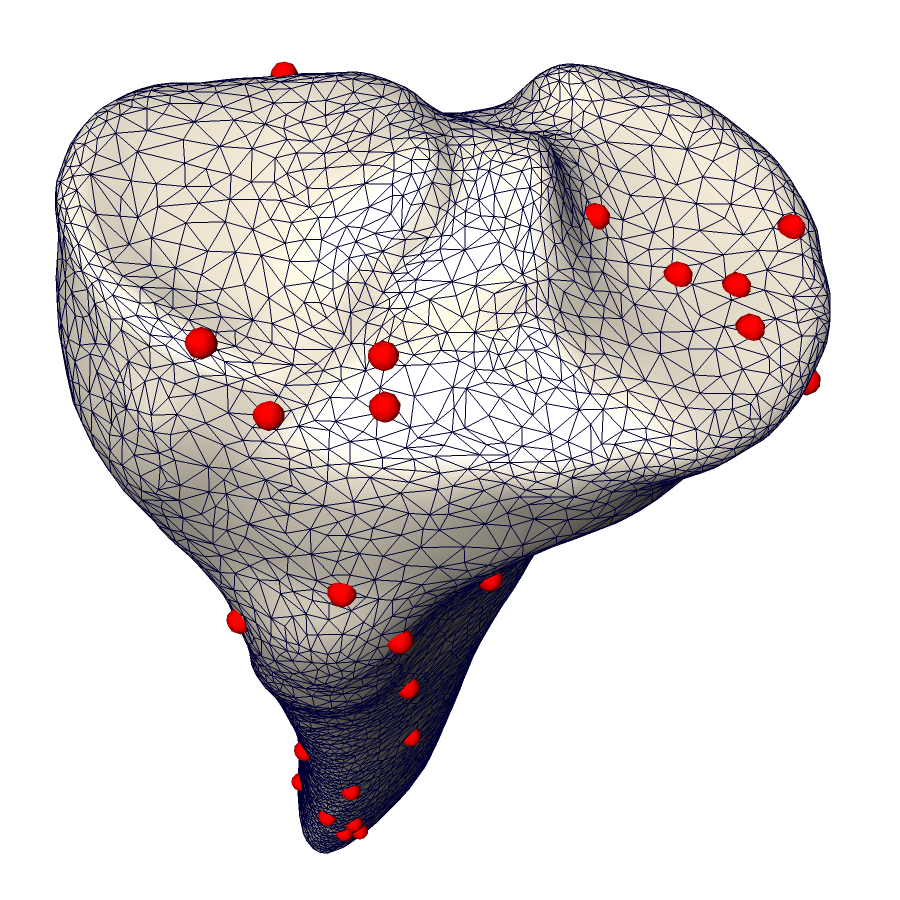}} \hfil 
       \subfigure[]{\includegraphics[scale=\tibiaScaleA]{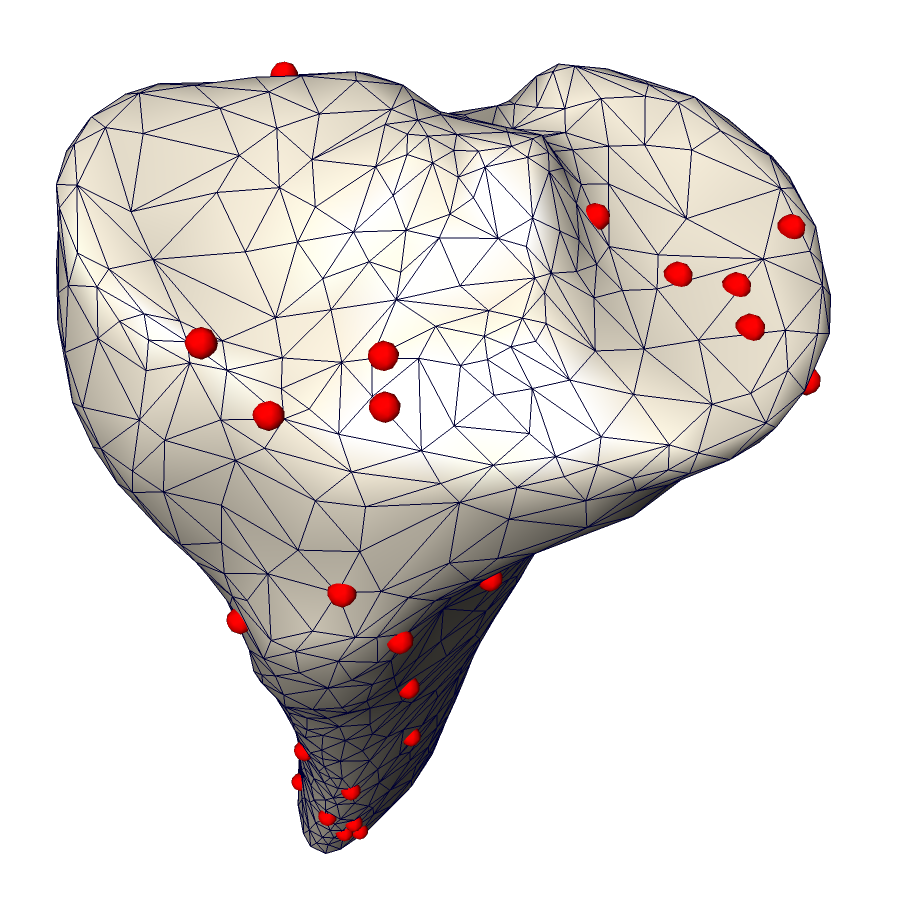}} 
     }
     \caption{Femur and tibia datasets. (a) Femur mean shape with $N=3800$ vertices. (b) Downsampled mean shape with $759$ vertices. (c) Tibia mean shape with $N=4701$ vertices. (d) Downsampled mean shape with $814$ vertices. For both bone models $P=36$ sparse points have been randomly drawn from the original surface according to the procedure described in section~\ref{drawRandomTrianglePts}.}
     \label{femur_tibia_models}
\end{figure}

For the femur and tibia datasets we assumed that the pose has already been normalised and we worked directly in the space of the SSM. 
In practice, this can for example be tackled in a similar manner as by \citet{Seim:2010autoseg}, who proposed an automated SSM-based knee bone segmentation, where initially the model is positioned inside 
the three dimensional CT or MR image via Generalised Hough Transform \citep{ballard1981generalizing}.

We present experiments for a PDM of the femur with $N=3800$ points (cf.~Fig.~\ref{femur_tibia_models}~(a)) and a PDM of the tibia with $N=4701$ points (cf.~Fig.~\ref{femur_tibia_models}~(c)). 
Additionally, we evaluated the ANISO method using the downsampled PDMs, denoted ANISO-ds, where only a subset of the original 
PDM vertices are used (cf.~Fig.~\ref{femur_tibia_models}~(b)~and~Fig.~\ref{femur_tibia_models}~(d)). Random sparse points are generated according to the procedure described in section~\ref{drawRandomTrianglePts}, 
where for each training shape 10 instances of sparse points $\mathcal{P}$ are sampled. In Fig.~\ref{femur_tibia_models} such random instances of $\mathcal{P}$ are shown for the mean shapes of both bones.
Summaries of the results are shown in Fig.~\ref{femurResultsSummary} for the femur and in Fig.~\ref{tibiaResultsSummary} for the tibia. 

\meanStdPlotsA{femur}{femur}{}{h!}
\meanStdPlotsA{tibia}{tibia}{}{h!}

It can be seen that for both bone PDMs if only $P=9$ points are available, the ICP methods perform slightly better than the ANISO method. 
Once more points become available, the ANISO method outperforms the ICP method.
Surprisingly, the ANISO-ds method, which uses a downsampled PDM, outperforms the ANISO method for $P=9$. 
We assume that this is because the original PDMs, comprising $N=3800$ vertices for the femur and $N=4701$ vertices for the tibia, contain fine local details that lead to an overfitting when reconstructing the surface from only $P=9$ points.
In contrast, the downsampled PDM contains less details that may impede the surface reconstruction. Moreover, the ANISO-ds method outperforms the ISO-ds method, which confirms our elaborations in Fig.~\ref{didex} on real data.
Moreover, for both bone PDMs, the AICP method performs very similar to ICP.

\subsection{Liver}\label{liver}

We carried out LAI and LOO experiments using a liver PDM with $N=4542$ points (cf.~Fig.~\ref{liverPdm}~(a)). 
Figure~\ref{liverPdm}~(b) shows the downsampled PDM. %
For each training shape $10$ instances of sparse points $\mathcal{P}$ are sampled. Summaries of the results are shown in Fig.~\ref{liverResultsSummary}. 
\newcommand{\liverscaleA}{.165}
\begin{figure}[h!]  
\centerline{
\subfigure[]{\includegraphics[scale=\liverscaleA]{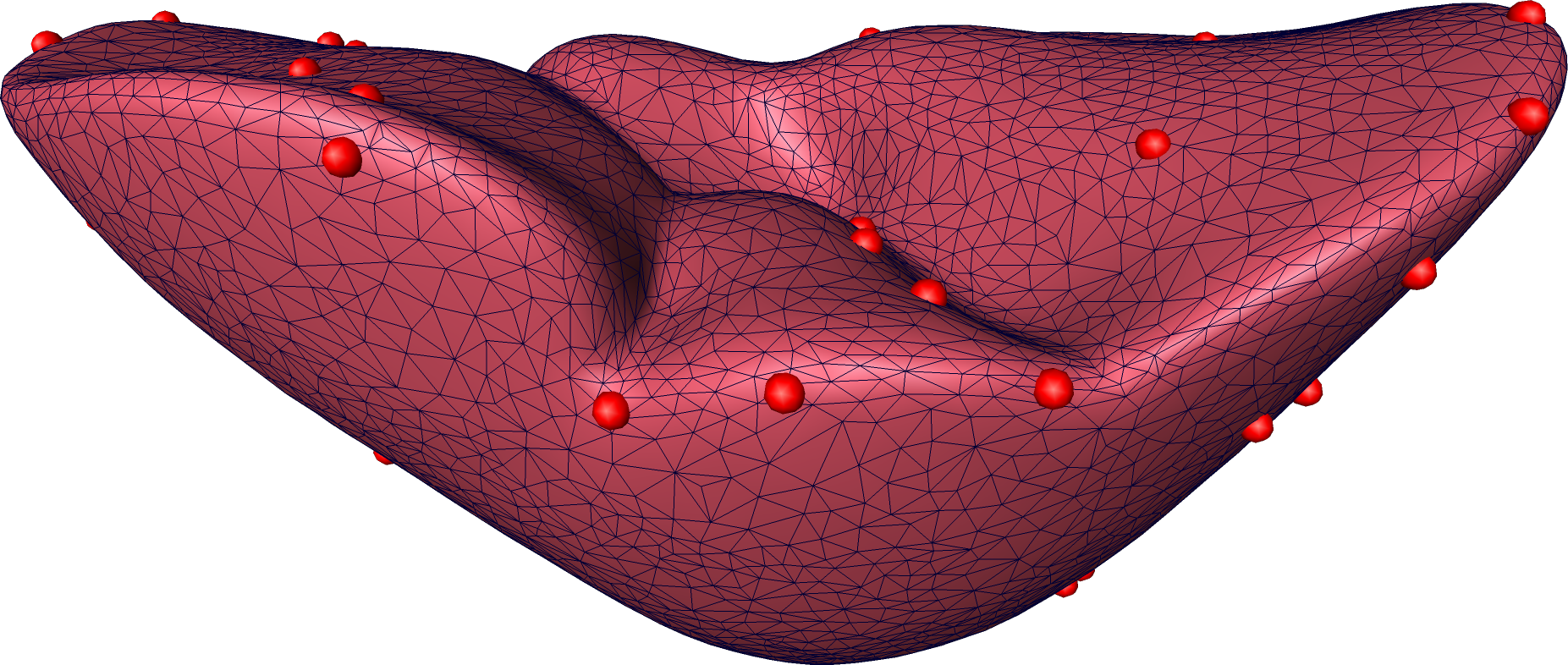}}}
\vspace{0.5mm}
\centerline{
\subfigure[]{\includegraphics[scale=\liverscaleA]{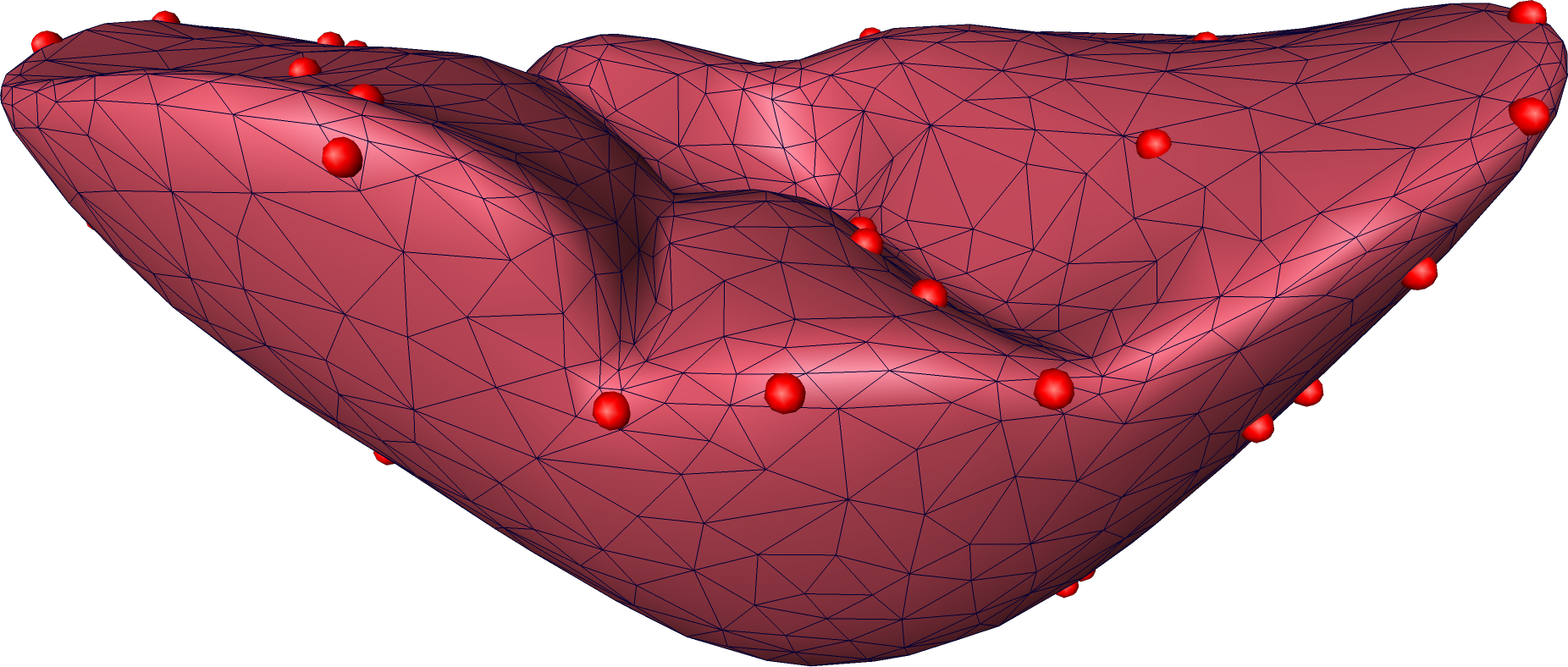}}
}
\caption{Liver dataset. (a) Mean shape with $N=4542$ vertices. (b) Downsampled mean shape with $908$ vertices. $P=36$ sparse points have been randomly drawn from the original surface according to the procedure described in section~\ref{drawRandomTrianglePts}.}
     \label{liverPdm}
\end{figure}
\meanStdPlotsA{liver}{liver}{}{b!}

For both settings, LAI and LOO, without any noise, i.e. $\tilde{\sigma} = 0$ cm, the anisotropic method outperforms both ICP methods with respect to the mean DSC regardless of how many random points have been sampled. 
Especially for $36$ and $90$ points the accuracy of the anisotropic method becomes increasingly superior compared to ICP.
Considering the random points disturbed by Gaussian noise with $\tilde{\sigma} = 1$ cm, the ICP method yields better results for $9$ and $18$ points. With noisy points, for the LAI and the LOO experiments at least 36 points seem to
be necessary for the anisotropic method to achieve results similar to the ICP methods with respect to the DSC. Similarly to the femur and tibia cases, the AICP method is on par with the ICP method.

\subsection{Hip}\label{hip}

We carried out LAI and LOO experiments using a hip PDM with $N=5603$ points (cf.~Fig.~\ref{hipPdm}~(a)). 
Figure~\ref{hipPdm}~(b) shows the downsampled PDM. %
\newcommand{\hipscaleA}{.16}
\begin{figure}[!b]  
\centerline{%
\subfigure[]{\includegraphics[scale=\hipscaleA]{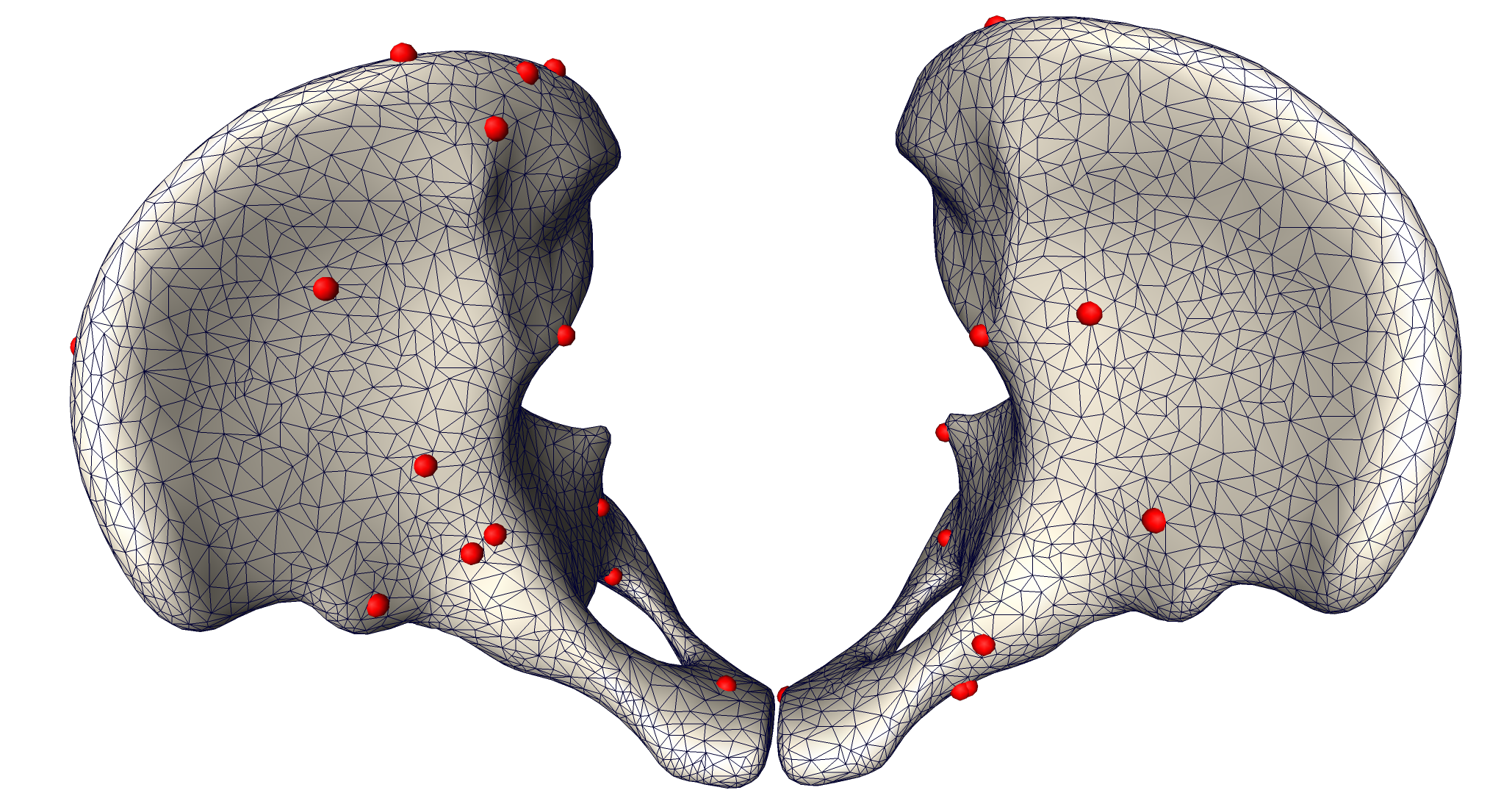}}%
} \centerline{
\subfigure[]{\includegraphics[scale=\hipscaleA]{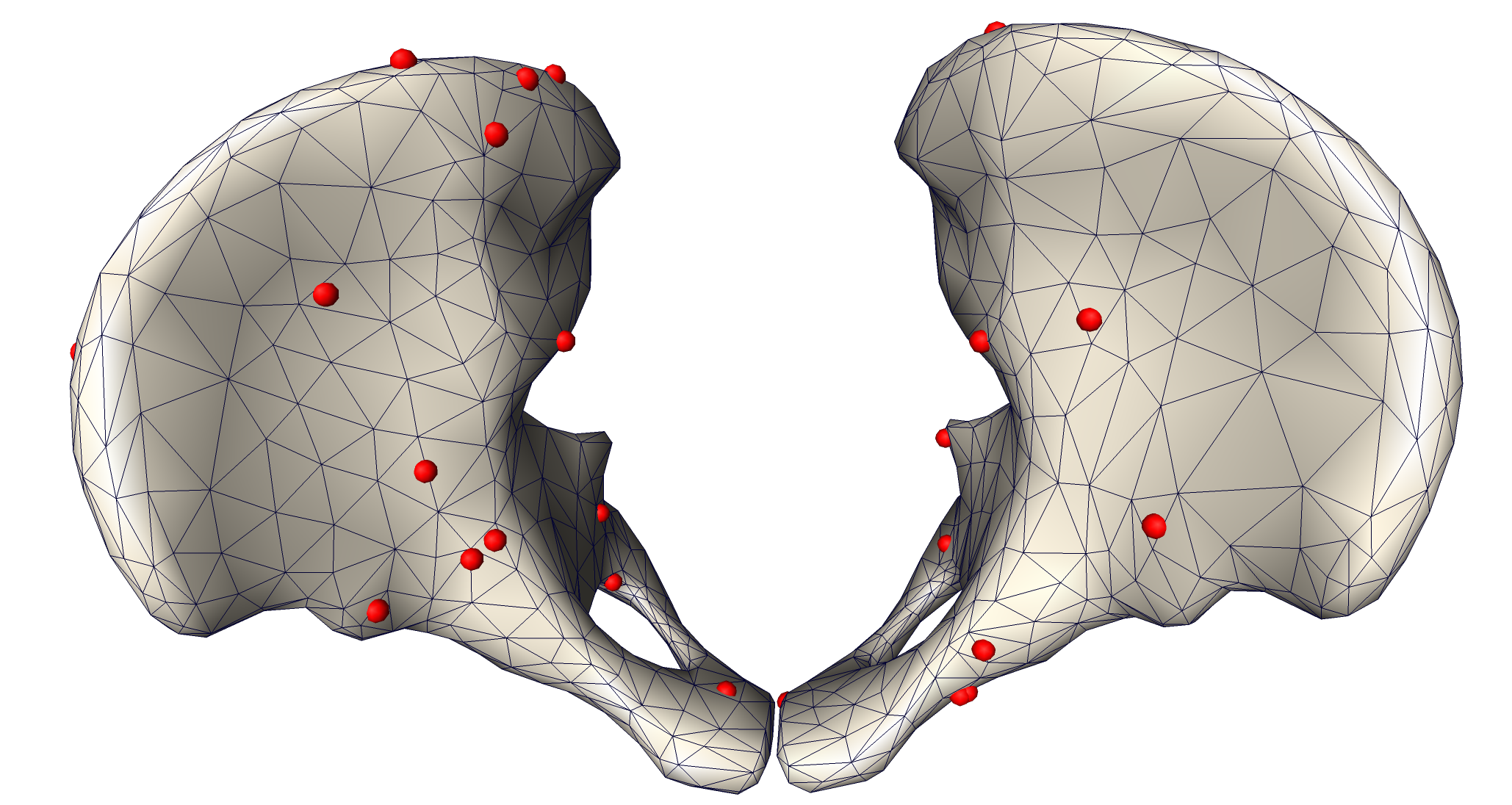}}%
}
     \caption{Hip dataset. (a) Mean shape with $N=5603$ vertices. (b) Downsampled mean shape with $1120$ vertices. $P=36$ sparse points have been randomly drawn from the original surface according to the procedure described in section~\ref{drawRandomTrianglePts}.}
     \label{hipPdm}
\end{figure}
For each training shape 10 instances of sparse points $\mathcal{P}$ are sampled.
Summaries of the results are shown in Fig.~\ref{hipResultsSummary}. 
For both settings, LAI and LOO, without any Gaussian disturbance, i.e. $\tilde{\sigma} = 0$ mm, the anisotropic method outperforms the ICP method with respect to the mean Dice Similarity Coefficient for more than $9$ points. Again, for the ANISO method the accuracy is increasing with more sampled points.
Considering the random points disturbed by Gaussian noise with $\tilde{\sigma} = 5$mm, the ICP method yields better results for $9$ points, whereas for $18$ points or more the ANISO method performs better. Comparing the ICP and the AICP approach, both methods are similar with respect to the DSC.

\meanStdPlotsA{hip}{hip}{}{h!}

\subsection{Brain Structures}\label{brainShapes}

In this experiment we consider a multi-object PDM that captures the inter-relation between multiple brain structures, namely \emph{Substantia Nigra \& Subthalamic Nucleus} (SN+STN, as compound object), \emph{Nucleus Ruber} (NR), \emph{Thalamus} (Th) and \emph{Putamen \& Globus Pallidus} (Put+GP, as compound object), where all structures are considered bilaterally. The mean of the PDM is shown in Fig.~\ref{brainPdm} (a). 

\newcommand{\brainscaleA}{.17}
\begin{figure*}[!h!]  
     \centerline{    
       \subfigure[]{\includegraphics[scale=\brainscaleA]{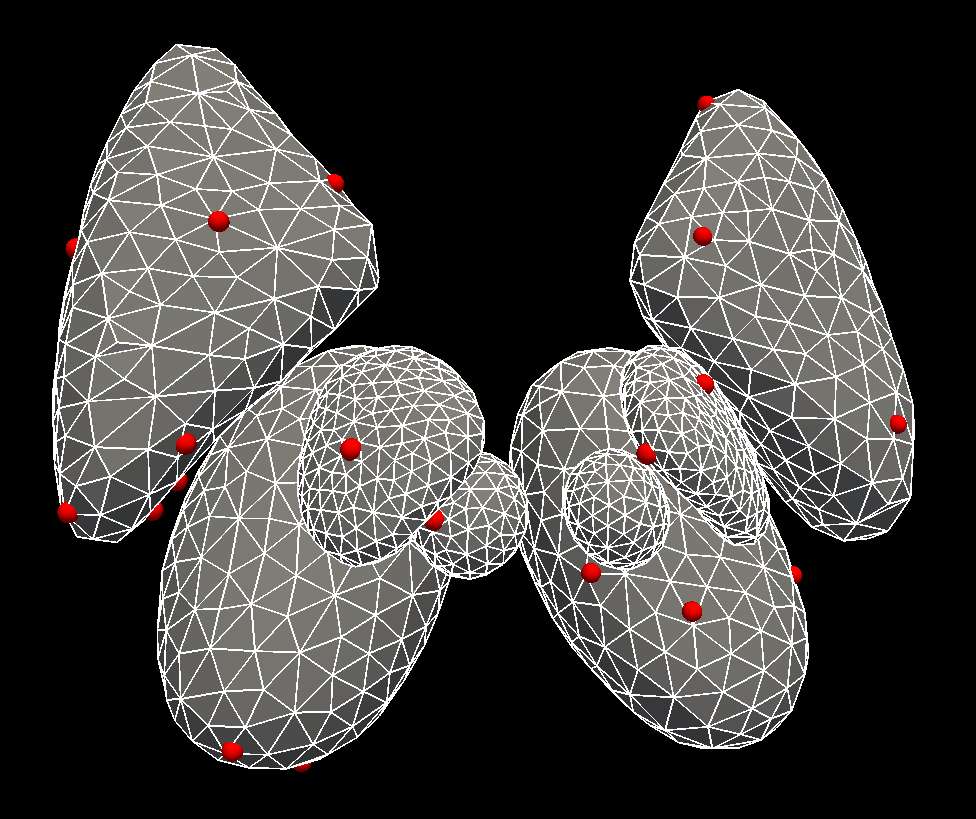}} \hfil   
       \subfigure[]{\includegraphics[scale=\brainscaleA]{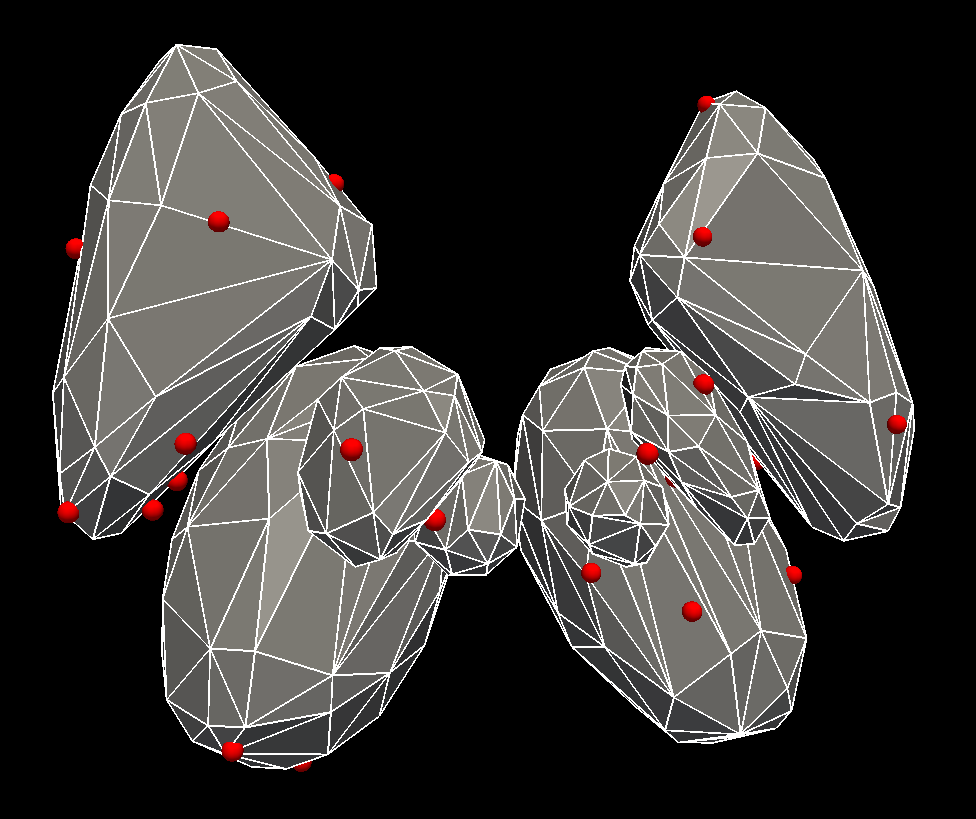}} \hfil
       \subfigure[]{\includegraphics[scale=\brainscaleA]{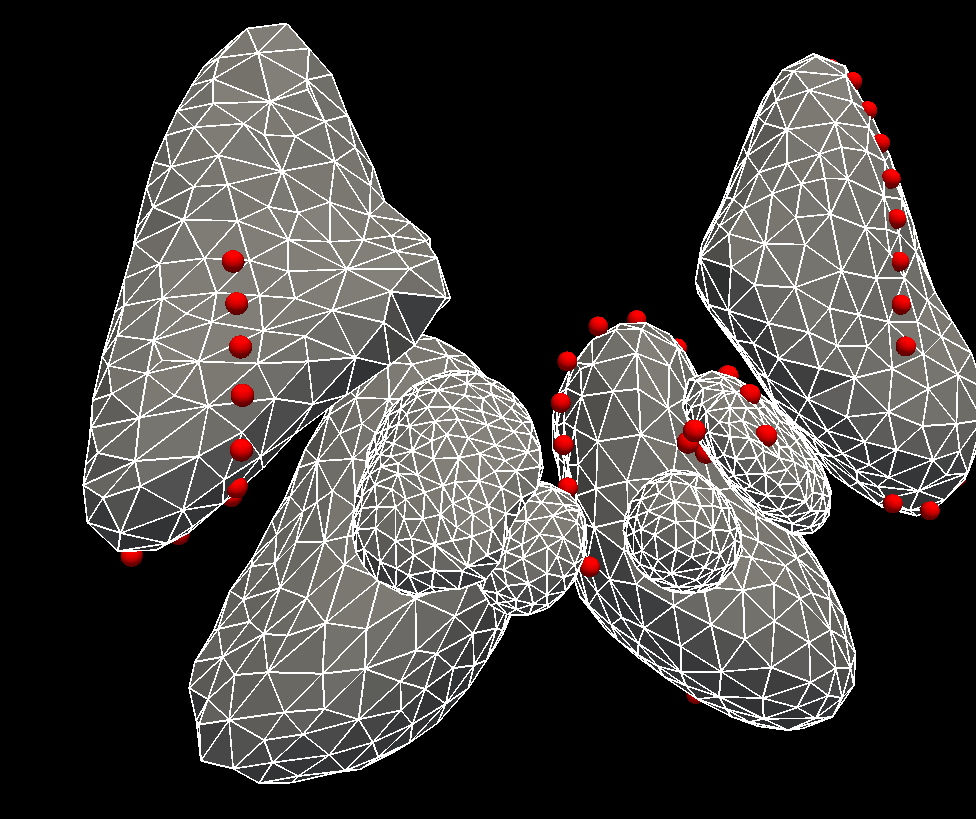}}      
     }
     \caption{Brain shapes dataset. (a) Mean shape with $N=1792$ vertices. (b) Downsampled mean shape with $371$ vertices. In (a,b) $P=36$ sparse points have been randomly drawn from the original surface according to the procedure described in section~\ref{drawRandomTrianglePts}. (c) A shape instance from the training set with partial contours.}
     \label{brainPdm}
\end{figure*}

The PDM is learned from multi-label segmentations that are all represented in a common coordinate system, the MNI ICBM 152 \citep{Fonov:2009bh} template space in our case (more details on the manual annotation and the establishment of correspondences can be found in our previous work \citep{bernard14bmt,Bernard:2016tv}). The alignment of the patient images into the MNI template space is conducted using the rigid image registration method FLIRT \citep{Jenkinson:2001wm}. Hence, thanks to this alignment, the orientation and position are already approximately normalised.
Consequently, for a new patient image that is to be segmented, a registration to the MNI template space is sufficient.

\subsubsection{Interactive Segmentation}\label{intSeg}
One interesting perspective of our presented shape-aware surface reconstruction method might be its integration into an interactive segmentation setting. This could be implemented by alternating between the user annotating object boundaries, and running our fitting method in order to reconstruct a surface from the user-input. 
As a first step into this direction, in addition to random sparse points, for the brain structure dataset we also evaluate a fitting to partial contours. We decided to focus on \emph{partial} contours instead of full contours since for some of the structures some regions may be difficult to delineate. In order to perform this evaluation we synthetically generated contours according to the procedure described in the supplementary material. In Fig.~\ref{brainPdm} (c) we show the sparse points that constitute these partial contours.
The main idea of the contour generation is as follows. First, we randomly select a 2D slice of the binary 3D segmentation image for a particular object. Next, from the 2D slice of the ground truth segmentation a subcontour of the entire boundary is randomly selected. Combining the chosen slice index with the subcontour leads to a planar 3D contour. Eventually, this contour is subsampled and the points are added to $\mathcal{P}$. Whilst our synthetic contour generation does not substitute a proper study involving user-drawn contours, we found that the resulting contours look plausible to be drawn by a human operator. For the generation of the partial contours we considered two settings, $\mathfrak{c}_1$ and $\mathfrak{c}_2$. For $\mathfrak{c}_1$, we have two contours in four of the eight brain structures, as shown in Fig.~\ref{brainPdm} (c), where the number of points ranges from $58$ to $106$, with a median of $80$. For $\mathfrak{c}_2$, we have a single contour for each of the eight brain structures, where the number of points ranges from $58$ to $81$, with a median of $68$.
 Note that when considering partial contours, for each $p_j \in \mathcal{P}$ we assume that it is known to which of the eight brain structures it belongs, which is used to constrain the E-step in our fitting methods (and the nearest-neighbour routine for the ICP methods).

In order to evaluate the robustness of our presented method with respect to noisy inputs, we also created noisy contours. For that, each partial contour is translated in the image plane by a random vector that has a zero-mean Gaussian distribution with covariance $\tilde{\sigma}^2 \matI_2$. Our motivation for using \emph{in-plane} translations is that when the user draws a contour in the image, the particular image plane is fixed and thus the only uncertainty occurs in-plane.

\subsubsection{Results} 

For each of the $K=17$ training shapes we sample $20$ instances of sparse points $\mathcal{P}$.
Following concepts introduced by \cite{Cootes:1995uz,Wang:1998bw,Wang:2000iw}, in the LOO experiments we increase the flexibility of the resulting shape model by extracting $M=96$ eigenvectors of a modified covariance matrix. In our case, we used the sum of the (scaled) covariance matrix $\matC$ and a Gaussian kernel with standard deviation of approximately $5$mm. We refer the interested reader to our previous work for details \citep{Bernard:2015wh}, where the method is referred to as \emph{kPCA}. We also demonstrated that this method is able to improve the generalisation ability of the PDM in this small training dataset comprising $K=17$ shapes.
Summaries of the results are shown in Fig.~\ref{brain_shapesResultsSummary}. 

\meanStdPlotsA{brain_shapes}{brain shape}{_kernelSVD_M-96}{h!}

As anticipated, the plots confirm that with respect to fitting accuracy an increasing number of measurements $P$ improves the results. Moreover, it can be seen that the ANISO method outperforms the ICP methods in all cases, where the standard deviation of ICP is much larger. In all cases, running \emph{any} fitting method is superior compared to simply using the mean shape.
Due to the simplicity of the ICP method, its runtime is much lower compared to the proposed fitting methods. However, by using the ANISO method with a downsampled PDM, the runtime can be reduced compared to the ANISO method, whilst still having superior fitting accuracy compared to ICP.

When using the method for interactive segmentation, the individual annotation of a moderate amount of random points, e.g.~$P=90$, is rather tedious and time-demanding. Thus, in the settings $\mathfrak{c}_1$ and $\mathfrak{c}_2$ we have evaluated the alternative of using points that can be derived from a very few number of partial contours. With that, one can obtain a reasonable number of points, cf.~section~\ref{intSeg}, with much less effort. Our results suggest that this is in general preferable over using a small amount of random points, say $P\leq18$. Nevertheless, the case of having $P=90$ random points outperformed the considered partial contours. We believe the reason is that 
the random points contain more diverse and scattered information compared to contours containing many correlated points.

\section{Conclusion and Outlook}\label{conclusion}
In this paper we have presented a methodology for a shape-aware surface reconstruction from sparse surface points. The proposed methodology is superior compared to the standard approach of ICP with respect to accuracy and robustness on a wide range of datasets.
In this method, the likely shape of the object that is to be reconstructed is captured by a PDM associated with a surface mesh. By interpreting the available sparse surface points as samples drawn from a GMM, the surface reconstruction task is cast as the maximisation of the posterior likelihood, which we tackle by variants of the EM algorithm. In order to achieve a surface-based fitting, we use a GMM with anisotropic covariance matrices, which are ``oriented'' by the surface normals at the PDM points. However, this results in a non-concave optimisation problem that needs to be solved in each M-step. We deal with this by maximising a concave approximation that considers the surface normals of the PDM computed from the previous value of the shape deformation parameter. 
As stated before, this approximation makes sense with the assumption that neighbour PDM vertices vary smoothly and the fact that surface normals are invariant to translations. We empirically demonstrated that finding a global maximum of this approximation leads to better results compared to finding a local optimum during the exact (non-concave) M-step. Moreover, our proposed concave approximation results in an algorithm that has the same time complexity as the isotropic fitting procedure.

The proposed surface reconstruction method deals exclusively with shape deformations. Thus, the normalisation of the pose must be solved a-priori in an application-dependent manner. In the example of the multi-object brain shape reconstruction we dealt with this issue by first performing rigid image registration in order to align the data into a common coordinate system.
Dealing with the limitation of not explicitly considering a rigid transformation in order to model the pose of the object is the next step for achieving an even broader applicability.
Whilst in principal one can formulate an analogous problem that considers the pose, the resulting problem is much more difficult to solve. This is because a simultaneous maximisation must be performed with respect to the rigid transformation and the shape deformation parameter. This is usually done iteratively, as in Active Shape Model search \citep{Cootes:1992uw}. With that, particular challenges to be dealt with are that the resulting surface reconstruction procedure would be much more sensitive to unwanted local optima as well as much slower. 

We have conducted an evaluation of the proposed algorithmic tools on a wide range of datasets in order to demonstrate their general applicability in the field of medical image analysis. For the evaluation we have considered a general noise model, i.e.~Gaussian noise that is independent for each point. In addition, for the contour case we also considered in-plane (Gaussian) noise. Since we focus on demonstrating the general applicability, a detailed evaluation of certain application-specific aspects (e.g. specific noise models) has not been studied in this paper. One interesting direction for future work is to consider outliers in the sparse points, which is relevant if the sparse points are automatically generated (e.g. using feature extraction methods). This could for example be tackled by integrating an additional uniform component into the mixture model \citep{Myronenko:2010wn}. Another approach is to use a RANSAC-like procedure \citep{Fischler:1981cv}.
In order to encourage the integration of our method into application-specific medical imaging workflows, we make our method publicly available\footnote{\url{https://github.com/fbernardpi/sparsePdmFitting}}. 
We expect that the public availability of the method will stimulate the commencement of interesting new research questions.
One such question may be which points are most useful for the reconstruction of surfaces.

\section*{Acknowledgement}
The authors gratefully acknowledge the financial support by the Fonds National de la Recherche, Luxembourg (6538106, 8864515, 9169303), by the German federal ministry of education and research (BMBF), grant no. 01EC1408B, and by the Einstein Center for Mathematics (ECMath), Berlin.

\appendix
\section{Gradient of $Q$ in~\eqref{qaniso}}\label{appendixGradient}
In the following, we derive the gradient of $Q$ w.r.t. $\vecalpha$, i.e.
\begin{align}
  \nabla_{\vecalpha} Q =
   \begin{bmatrix}
    \frac{\partial Q}{\partial \vecalpha_m}
  \end{bmatrix}_m \,.
\end{align}
For brevity, we write $\partial \cdot$ to denote the partial derivative $\frac{\partial \cdot}{\partial \vecalpha_m}$ w.r.t. $\vecalpha_m$, where the dependence on $m$ is implicit.
First, we note that the cross-product $u \times v$ of two vectors $u,v \in \R^3$ can be written as the matrix multiplication $[u]_\times v$, where the operator $[\cdot]_\times: \R^3 \rightarrow \R^{3 \times 3}$ creates a skew-symmetric matrix from its input vector by
\begin{align}\label{skewsym}
  [\begin{pmatrix}
    u_1\\u_2\\u_3
  \end{pmatrix}]_\times := \begin{bmatrix}
    0 & -u_3 & u_2 \\ u_3 & 0 & -u_1 \\ -u_2 & u_1 & 0
  \end{bmatrix}\,.
\end{align}
Introducing
\begin{align}\label{b}
  b_i(\vecalpha) := (y_{i_2}(\vecalpha) - y_i(\vecalpha)) \times (y_{i_3}(\vecalpha) - y_i(\vecalpha)) \,,
\end{align}
we can write $n_i(\vecalpha) = \frac{b_i(\vecalpha)}{\|b_i(\vecalpha) \|}$. Now, by representing the cross product in \eqref{b} as a matrix product with the notation from \eqref{skewsym}, and by using the product rule, the partial derivative of $b_i(\vecalpha)$ is given by
\begin{align}
  \partial b_i(\vecalpha) = &~ \{\partial [y_{i_2}(\vecalpha) - y_i(\vecalpha)]_\times\} (y_{i_3} (\vecalpha)-y_i(\vecalpha)) \label{gradq} \\
  & ~+[y_{i_2}(\vecalpha) - y_i(\vecalpha)]_\times \{\partial  (y_{i_3} (\vecalpha)-y_i(\vecalpha))\} \nonumber\\
  = & ~[\matPhi_{i_2,m} - \matPhi_{i,m}]_\times (y_{i_3} (\vecalpha)-y_i(\vecalpha)) \\
  & ~+[y_{i_2}(\vecalpha) - y_i(\vecalpha)]_\times (\matPhi_{i_3,m} - \matPhi_{i,m}) \nonumber\,.
\end{align}
Moreover,
\begin{align}
  \partial \|b_i(\vecalpha)\| = \frac{b^T_i(\vecalpha) \{\partial b_i(\vecalpha)\} }{\|b_i(\vecalpha)\|}\,.
\end{align}
By using the quotient rule, the partial derivative of $n_i(\vecalpha)$ is given by
\begin{align}
  \partial n_i(\vecalpha) = &~ \frac{\|b_i(\vecalpha)\| \{\partial b_i(\vecalpha)\} - b_i(\vecalpha) \{\partial\|b_i(\vecalpha)\|\}}{ \|b_i(\vecalpha)\|^2}\,.
\end{align}
Using $\partial n_i(\vecalpha)$, we can write
\begin{align}
  \partial \matW_i(\vecalpha) = (\eta{-}1)(\{\partial n_i(\vecalpha)\}n_i^T(\vecalpha) + n_i(\vecalpha) \{\partial n_i^T(\vecalpha)\}) \,.
\end{align}
Now, given the expression for $\partial \matW_i(\vecalpha)$, we can finally compute the partial derivative of $Q$ w.r.t. $\vecalpha_m$, which is
\begin{align}
  &\partial Q(\vecalpha,\sigma, \vecalpha^{(n)}, \sigma^{(n)}) =  \\
  & \qquad -(\Lambda^{-1})_{m,:}\vecalpha  - \frac{1}{2\sigma^2} \sum_{i,j} \prob(i \vert p_j, \vecalpha^{(n)}, \sigma^{(n)}) \cdot \nonumber\\
  &\qquad\qquad\qquad[(p_j -y_i(\vecalpha))^T \{\partial \matW_i(\vecalpha)\}(p_j -y_i(\vecalpha)) - \nonumber\\
  &\qquad\qquad \qquad\quad2 \matPhi_{i,m}^T \matW_i(\vecalpha) (p_j - y_i(\vecalpha))] \nonumber \,.
\end{align}

\section*{References}
\bibliography{refs}

\end{document}